\documentclass[11pt]{article}

\usepackage[preprint]{acl}

\usepackage{times}
\usepackage{latexsym}
\usepackage[utf8]{inputenc}
\usepackage{csquotes}
\usepackage{graphicx}

\usepackage[T1]{fontenc}

\usepackage[utf8]{inputenc}

\usepackage{microtype}

\usepackage{inconsolata}

\usepackage{graphicx}
\usepackage{booktabs}
\usepackage{multicol}
\usepackage{multirow}
\usepackage{xcolor}
\usepackage[table]{xcolor}
\usepackage{algorithm}
\usepackage{algpseudocode}
\usepackage{makecell}
\usepackage{amsmath}
\usepackage{amssymb}
\usepackage{tcolorbox}
\tcbuselibrary{breakable}

\newtcolorbox{promptbox}{
  breakable,
  colback=gray!5,
  colframe=gray!40,
  boxrule=0.5pt,
  arc=2pt,
  left=6pt,
  right=6pt,
  top=6pt,
  bottom=6pt,
  fontupper=\ttfamily\small
}
\title{Instruction Quality Matters: \\
Refining Instructions for Effective Preference Learning}

\author{%
  Seohyeong Lee$^{1}$\;\;
  Hwaran Lee$^{1,}$\thanks{\; Corresponding author}\;\;
  Buru Chang$^{2,}$\footnotemark[1]\\
  \textsuperscript{1}Sogang University\;\;
  \textsuperscript{2}Korea University\\
  \texttt{\{seohyeong, hwaranlee\}@sogang.ac.kr} \;\;
  \texttt{buru\_chang@korea.ac.kr}
}

\begin{document}
\maketitle
\begin{abstract}\label{sec:0_abstract}
Preference learning optimizes models using response pairs, yet the informativeness of these pairs is fundamentally shaped by the instructions from which they are generated.
We identify instruction quality as a hidden bottleneck in preference learning: low-quality or ambiguous instructions restrict the response-quality distribution, limiting strong chosen responses and weakening preference signals.
Through Best- and Worst-of-$N$ analyses, we show that instruction quality constrains both the ceiling and floor of sampled response quality.
Motivated by this observation, we introduce an instruction-refinement pipeline that selects weak instructions using reward signals and revises them with rubric-guided LLM feedback, improving preference data without discarding examples.
Across offline and online preference learning settings, experiments on multiple models and benchmarks show broad alignment improvements over original data and alternative data-improvement strategies.
Further analyses indicate that instruction refinement raises achievable response quality and complements response-centric preference data curation.
Overall, instruction quality emerges as a key factor governing how informative preference signals are formed for LLM alignment.
Code is available at: \url{https://github.com/01choco/instruction-refinement/}
\end{abstract}
\section{Introduction}\label{sec:1_introduction}
\textit{Preference learning}~\cite{christiano2017deep} is a central paradigm for aligning Large Language Models (LLMs) with human preferences.
Despite advances in optimization methods~\cite{meng2024simpo, liu2024deepseek}, its effectiveness remains highly dependent on the quality of preference data~\cite{zhou2023lima}.
Existing work has primarily examined preference data quality through response quality and annotation reliability, showing that weak responses or noisy labels can substantially degrade alignment~\cite{pan2025what, lee2025dataset, chowdhuryprovably, gao2024impact}.
This view, however, treats the instruction as a fixed input, overlooking its role in shaping preference signals.
\begin{figure}[!t]
    \centering
    \includegraphics[width=\linewidth]{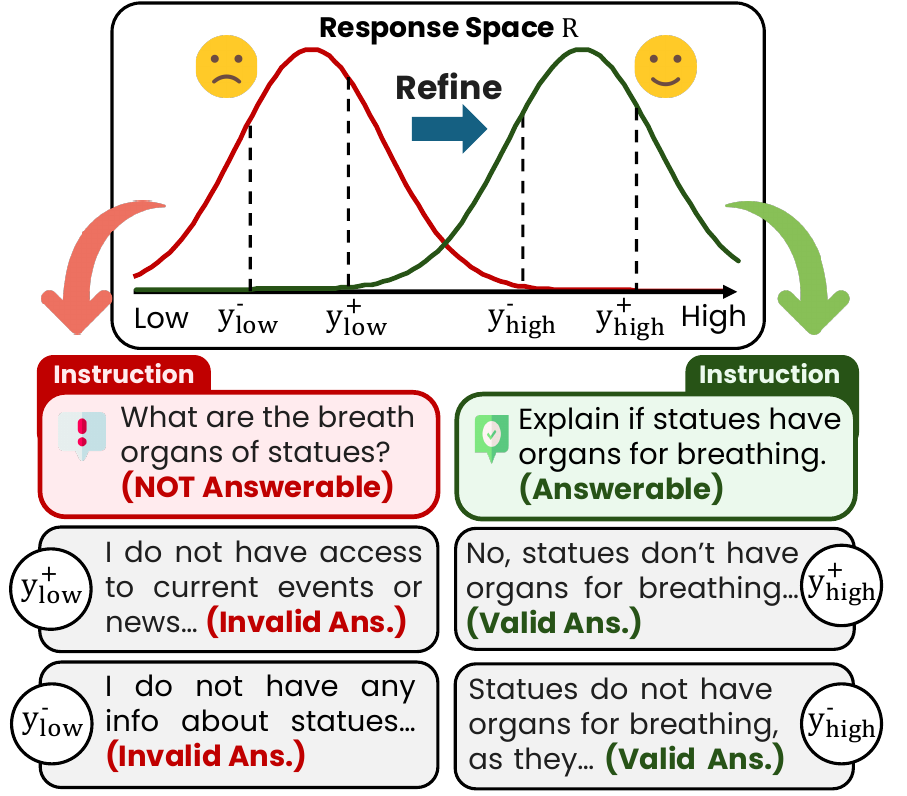}
    \caption{
        Data example from the UltraFeedback~\cite{cui2024ultrafeedback}.
        Ambiguous instructions can produce invalid responses, leading to weak preference learning signals.
        Refining such instructions yields high-quality outputs that provide stronger signals.
    }
    \label{fig:5_case}
    \vspace{-1em}
\end{figure}

Preference pairs are sampled from a response distribution induced by the instruction.
When an instruction is ambiguous, incomplete, or underspecified, this distribution may contain few high-quality responses and many trivial failures, making the resulting preference data less informative. 
As illustrated in Figure~\ref{fig:5_case}, such cases appear in preference datasets and can induce invalid or low-quality responses.
Because this candidate response pool is constrained upstream, response refinement or filtering can only partially mitigate the problem.

We first test this upstream effect by analyzing the response distribution from which preference pairs are constructed.
Using Best-of-$N$ and Worst-of-$N$ analyses, which approximate the response-quality ceiling and floor under a fixed sampling budget, we examine how instruction quality shapes sampled responses.
We find clear gaps between low- and high-quality instructions: high-quality instructions improve not only the best sampled responses but also the worst sampled responses.
These gaps indicate that low-quality instructions restrict useful candidate coverage, reducing both strong positive examples and informative preference contrasts for DPO-style optimization (see Section~\ref{subsec:3_motivating_experiments} and Figure~\ref{fig:1_plot_sample}).
Thus, while preference learning depends on response quality~\cite{pan2025what}, the response quality available for learning can itself be constrained by instruction quality.

To mitigate these upstream effects on preference-pair construction, we refine instructions before generating candidate responses.
Our pipeline selects weak instructions using reward signals and revises them with rubric-guided LLM feedback, improving preference data without discarding examples.
Rather than treating refinement as generic rewriting, the rubrics provide structured diagnostic feedback that guides revisions toward instructions that elicit reliable and comparable candidate responses.
Unlike inference-time prompt engineering, our approach targets the training data from which preference-learning signals are derived.

We evaluate instruction refinement on UltraFeedback~\cite{cui2024ultrafeedback} across offline and online preference learning.
In offline experiments, refinement improves alignment across multiple backbones, objectives, and benchmarks, including MT-Bench~\citep{zheng2023judging}, Evol-Instruct~\citep{xu2023wizardlm}, and AlpacaEval~\cite{alpaca_eval}.
In online experiments, even one refinement step per data-generation round improves iterative preference-learning pipelines such as SPA~\cite{kim2025spread}.
Across settings, refinement yields broad alignment gains of up to 8\%p, indicating that instruction quality remains a limiting factor even as models and preference data co-evolve.

Further analyses show that refinement improves instruction-quality scores and response rewards, raises the Best-of-$N$ response-quality ceiling, and remains robust across threshold choices.
A rubric ablation further shows that rubric-guided feedback outperforms generic instruction rewriting, indicating that structured diagnostic feedback is important for effective refinement.
These results support the view that instruction refinement improves preference learning by making the candidate response pool more informative.

Our contributions are summarized as follows:
\begin{itemize}
    \item We identify instruction quality as an upstream factor in preference-signal formation and show that it shapes response-quality ceilings and floors, affecting the informativeness of pairwise preference data.
    \item We introduce an instruction refinement pipeline that improves weak instructions using reward signals and rubric-guided LLM feedback, without discarding data.
    \item We validate the approach across offline and online preference learning, showing broad alignment gains and analyses that support its effect on response-quality distributions.
\end{itemize}

\section{Related Work}
\subsection{Preference Data Curation}
\noindent
Prior work has recognized the importance of data quality in preference learning and proposed methods for improving preference pairs in both offline and online settings.
In offline preference learning, existing approaches typically identify informative or reliable preference pairs based on response-level or pair-level signals, such as reward margins~\citep{deng2025less}, response quality~\citep{lee2025dataset}, or annotation consistency~\citep{lee2025preference}.
In online preference learning, prior work improves alignment by iteratively generating responses, constructing preference pairs, and updating the model while reducing noise in preference labels or refining response quality~\citep{kim2025spread, pan2025what, dong2025selfboosting}.
In contrast, these methods largely treat instructions as fixed inputs, whereas we study instruction quality as an upstream factor in preference data informativeness.


\begin{figure*}[!ht]
    \centering
    \includegraphics[width=\textwidth]{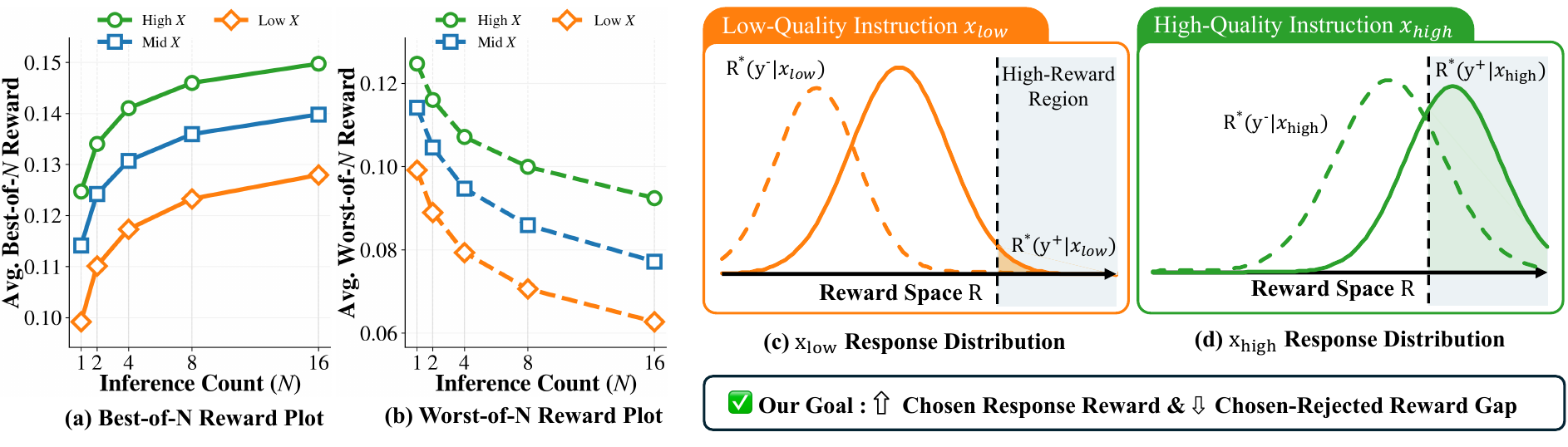}
    \caption{
    Instruction quality shapes response distributions and preference-pair informativeness.
    \textbf{Left}: (a) Best-of-$N$ reward shows that high-quality instructions raise the response-quality ceiling.
    (b) Worst-of-$N$ reward shows that they also raise the response-quality floor.
    \textbf{Right}: 
    (c): Reward score distributions of chosen $y^+$ and rejected $y^-$ responses paired with low-quality instructions $x_{low}$. (d): Those of high-quality instructions $x_{high}$. 
    }
    \label{fig:1_plot_sample}
    \vspace{-1em}
\end{figure*}

\subsection{Preference Data Refinement}

\noindent
Our work is methodologically related to self-refinement and broader data-improvement methods.
Self-Refine~\citep{madaan2023self}, Reflexion~\citep{shinn2023reflexion}, and recent refinement pipelines~\citep{cayir2025refine} use natural language feedback, reflection, or LLM-based judges to improve responses, agent behavior, or dataset quality.
However, these methods primarily refine responses while leaving instructions fixed.

Regarding instruction-based selection methods, R.I.P.~\citep{yu2025rip} shows that prompt quality affects model training by selecting high-quality prompts for instruction tuning.
\citet{li2024quantity} select instruction-tuning examples based on instruction-following difficulty.
In contrast, we focus on preference data curation by refining instructions for preference learning to elicit clearer preference signals.
This perspective is also related to recent work showing that reward models and instruction-tuned models can be sensitive to input or instruction variations~\citep{wu2025rewordbench, yan2024contrastive}.
While these studies target model robustness, we refine low-quality instructions to construct more informative preference data.
\section{Instruction Quality and Refinement}
\subsection{Motivating Analysis: Instruction Quality Shapes Response Distributions}
\label{subsec:3_motivating_experiments}

Preference pairs are constructed from instruction-conditioned candidate responses.
Ambiguous or underspecified instructions may yield few high-quality candidates and many trivial failures, weakening preference data before optimization begins.

To examine this effect, we first conduct Best-of-$N$ and Worst-of-$N$ analyses on 1.5K UltraFeedback~\cite{cui2024ultrafeedback} instructions, partitioned into High, Mid, and Low quality groups using rubric-based instruction scores.
For each instruction, we generate multiple responses with LLaMA3-8B and score them with a reward model.
Best-of-$N$ approximates the response-quality ceiling under a fixed sampling budget, while Worst-of-$N$ approximates the response-quality floor.

We use reward-model scores as explicit proxies for response quality, rather than as the rewards directly optimized by DPO.
DPO instead induces an implicit reward
\(r_\theta(x,y)=\beta \log \frac{\pi_\theta(y|x)}{\pi_{\mathrm{ref}}(y|x)}\).
Therefore, our interpretation assumes that explicit reward-model scores preserve the relevant ordering and margin trends of the implicit rewards that emerge during preference optimization.

As shown in Figure~\ref{fig:1_plot_sample}, high-quality instructions consistently achieve higher Best-of-$N$ and Worst-of-$N$ rewards than low-quality instructions.
This indicates that low-quality instructions both lower the response-quality ceiling and weaken the response-quality floor; they reduce access to strong candidate responses while increasing the prevalence of trivial low-quality responses.

We next connect these ceiling and floor effects to preference-gradient informativeness.
For a preference tuple \((x,y^+,y^-)\), the DPO gradient can be written using the preference margin \(u_\theta\) and likelihood contrast \(d_\theta\):
\[
\begin{aligned}
\nabla_\theta \mathcal{L}_{\mathrm{DPO}}
&=
-\beta \, \mathbb{E}
\left[
\sigma(-u_\theta)\, d_\theta
\right], \\
d_\theta
&=
\nabla_\theta \log \pi_\theta(y^+|x)
-
\nabla_\theta \log \pi_\theta(y^-|x), \\
u_\theta
&=
\beta
\left[
\log \frac{\pi_\theta(y^+|x)}{\pi_{\mathrm{ref}}(y^+|x)}
-
\log \frac{\pi_\theta(y^-|x)}{\pi_{\mathrm{ref}}(y^-|x)}
\right].
\end{aligned}
\]
Here, \(d_\theta\) encodes the gradient direction favoring \(y^+\) over \(y^-\), while \(\sigma(-u_\theta)\) 
decays as the current policy $\pi_\theta$ already separates the pair (i.e., as $u_\theta$ grows during training).
A higher response-quality ceiling increases the chance that \(y^+\) provides a high-quality target for optimization, enabling the positive term in \(d_\theta\) to learn from stronger targets.
A higher response-quality floor makes rejected responses less degenerate.
When \(y^-\) is trivially poor, the chosen--rejected contrast can become overly easy: as training progresses, such pairs are likely to attain a large implicit margin \(u_\theta\), leading to a small DPO gradient weight \(\sigma(-u_\theta)\).
By reducing degenerate negatives, a higher floor helps preserve nontrivial preference contrasts, making preference pairs informative for optimization~\citep{houliston2024uncertainty}.
Thus, ceiling and floor improvements are complementary: the former provides stronger positive examples, while the latter reduces degenerate negatives without eliminating meaningful chosen--rejected gaps.
This motivates instruction refinement before candidate generation; Appendix~\ref{subsec:a_1_motivating} provides a DPO-based interpretation.

\subsection{Refining Instructions for Preference Data}
\label{sec:refinement_algorithm}

We view a preference dataset as tuples \((X,Y^+,Y^-)\), where \(X\) is an instruction and \(Y^+,Y^-\) are chosen and rejected responses.
Instead of refining responses after generation, we refine \(X\) before constructing preference pairs, aiming to improve the candidate response pool from which preference signals are formed.

\begin{algorithm}[t]
\small
\caption{Instruction Refinement Loop}
\label{alg:refinement}
\begin{algorithmic}[1]
\Require Instruction Set $\{\mathcal{X}\}$, Reward model $R(\cdot)$, Threshold $\tau$, Max iterations $T$
\Ensure Refined On-Policy Dataset $\mathcal{D} = (X', Y^+, Y^-)$
\State Initialize $\mathcal{D}' \gets \emptyset$
\For{Each instruction $X \in \{\mathcal{X}\}$}
  \State Generate $(Y_1, Y_2)$ from the target model given $X$
  \State Compute rewards $R_1 \gets R(X, Y_1)$, $R_2 \gets R(X, Y_2)$
  \State $t \gets 0$
  \While{$\min(R_1, R_2) < \tau$ \textbf{and} $t < T$}
    \State Generate rubric-based feedback for $X$
    \State Refine instruction $X \gets X'$
    \State Regenerate $(Y_1, Y_2)$ and recompute $(R_1, R_2)$
    \State $t \gets t + 1$
  \EndWhile
  \State Add refined tuple $(X, Y^+, Y^-)$ to $\mathcal{D}$
\EndFor
\State \Return $\mathcal{D}$
\end{algorithmic}
\end{algorithm}

Algorithm~\ref{alg:refinement} summarizes the refinement procedure.
For each instruction \(X\), we generate on-policy responses \((Y_1,Y_2)\) with the target model and compute reward scores \(R_i=R(X,Y_i)\).
If \(\min(R_1,R_2)<\tau\), where \(\tau\) is a minimum-reward threshold, we select \(X\) for refinement, treating the low minimum reward as a proxy for lower-tail failures induced by the instruction.
Selected instructions are revised using rubric-based feedback; after each revision, responses and rewards are regenerated.
The loop repeats until \(\min(R_1,R_2)\ge\tau\) or the maximum number of iterations is reached\footnote{\textbf{API call cost.}
Instruction refinement adds two API calls per sample for feedback and revision, costing about \$0.0026 per sample in our setup.
While manageable in our experiments, this overhead may become substantial at larger scales.
}.

This feedback--refinement loop follows the spirit of Self-Refine~\cite{madaan2023self}, but applies refinement to instructions in preference data rather than to generated responses, improving training data without discarding examples.

\subsection{Evaluation and Refinement Criteria}
\label{subsec:3_2_eval_ref_instruction}

We use rubric-based evaluation as structured diagnostic feedback for instruction refinement in preference data.
Reward scores identify instructions that induce weak candidates, while rubrics diagnose why they fail and how they should be revised.
Specifically, the rubrics assess whether an instruction is likely to elicit reliable and comparable candidate responses for preference learning, covering clarity, specificity, completeness, safety, answerability, conciseness, and format consistency.
Detailed definitions, motivations, and supporting studies are provided in Appendix~\ref{appendix:rubric_design}.

\noindent\textbf{Scoring and Feedback Generation.}
For each instruction \(X\), an \texttt{LLM-as-a-judge}~\cite{zheng2023judging} evaluates all rubrics and produces structured feedback \(\mathcal{F}\).
For each rubric, the evaluator outputs a 1--5 score and a short natural-language explanation.
The scores provide a quantitative estimate of instruction quality, while the explanations identify the instruction-level deficiencies that should be revised.
We average the rubric scores to obtain an overall instruction-quality score.

\noindent\textbf{Refinement.}
The refiner takes the original instruction \(X\) and feedback \(\mathcal{F}\) as input, and outputs a revised instruction \(X'\) that addresses the diagnosed deficiencies.
The full feedback and refinement prompts are provided in Appendix~\ref{A_6_prompt}

\begin{table*}[!ht]
\centering
\resizebox{0.9\textwidth}{!}{
\begin{tabular}{c|c|cc|cc|cc|cc|cc|cc}
\toprule
\multicolumn{2}{c|}{\textbf{Method}}& \multicolumn{6}{c|}{DPO} & \multicolumn{6}{c}{SimPO} \\
\midrule
\multirow{2}{*}{\textbf{Backbone}} & \multirow{2}{*}{\textbf{Data Type}} & \multicolumn{2}{c|}{MT-Bench} & \multicolumn{2}{c|}{Evol-Instruct} &  \multicolumn{2}{c|}{AlpacaEval 2.0} & \multicolumn{2}{c|}{MT-Bench} & \multicolumn{2}{c|}{Evol-Instruct} & \multicolumn{2}{c}{AlpacaEval 2.0} \\
& & WR & Score & WR & Score & WR & \textbf{LC WR} & WR & Score & WR & Score & WR & \textbf{LC WR} \\
\midrule
\multirow{2}{*}{Llama3-8b}
& SFT & 27.8 & 4.61 & 23.4 & 5.80 & 3.95 & 7.49 & 27.8 & 4.61 & 23.4 & 5.80 & 3.95 & 7.49 \\
\cmidrule(lr){2-14}

& Original & 39.7 & 4.91 & 33.3 & 6.15 & 5.91 & 11.90 & 47.5 & 5.26 & 40.8 & 6.53 & 8.65 & 15.90 \\
\rowcolor{gray!10} \cellcolor{white} & \cellcolor{white} Refined & \textbf{47.8} & \textbf{5.38} & \textbf{35.8} & \textbf{6.39} & \textbf{8.86} & \textbf{18.78} & \textbf{48.1} & \textbf{5.56} & \textbf{44.5} & \textbf{6.62} & \textbf{9.19} & \textbf{20.30} \\

\midrule
\multirow{2}{*}{Mistral-7b}
& SFT & 21.6 & 4.04 & 16.3 & 4.87 & 4.30 & 8.25 & 21.6 & 4.04 & 16.3 & 4.87 & 4.30 & 8.25 \\
\cmidrule(lr){2-14}
& Original & 35.6 & \textbf{5.03} & 31.0 & \textbf{6.35} & 7.80 & 12.95 & 24.1 & 4.18 & 12.4 & 5.56 & 8.09 & 16.83\\
\rowcolor{gray!10} \cellcolor{white} & \cellcolor{white} Refined & \textbf{43.4} & 4.96 & \textbf{33.3} & 6.24 & \textbf{8.21} & \textbf{14.60} & \textbf{28.8} & \textbf{4.43} & \textbf{19.5} & \textbf{5.72} & \textbf{9.18} & \textbf{17.40} \\

\bottomrule
\end{tabular}
}
\vspace*{-0.5em}
\caption{
Evaluation results on MT-Bench, Evol-Instruct, and AlpacaEval.
For MT-Bench and Evol-Instruct, each cell reports the pairwise win rate and single-answer score; for AlpacaEval, each cell reports the win rate and length-controlled win rate.
The highest values for each backbone are highlighted in \textbf{bold}.
}
\label{tab:1_offline}
\vspace*{-0.5em}
\end{table*}

\begin{table}[ht]
\centering
\resizebox{\columnwidth}{!}{
\begin{tabular}{l|cc|cc|cc}
\toprule
\textbf{Method}
& \multicolumn{2}{c|}{MT-Bench} 
& \multicolumn{2}{c|}{Evol-Instruct} 
& \multicolumn{2}{c}{AlpacaEval} \\
& \textbf{WR} & \textbf{Score} 
& \textbf{WR} & \textbf{Score} 
& \textbf{WR} & \textbf{LC WR} \\
\midrule
\textbf{SFT}
& 27.8 & 4.61 
& 23.4 & 5.80 
& 3.55 & 7.15 \\
\textbf{DPO} (Original Dataset)
& 39.7 & 4.91 
& 33.3 & 6.15 
& 5.91 & 11.90 \\
\midrule
\textbf{Prompt Engineering} & & & & & & \\
\quad \textit{CoT Prompting}~\citep{wei2022chain}
& 30.6 & 4.68 
& 28.2 & 5.98 
& 4.82 & 9.42 \\
\quad \textit{Paraphrasing}~\citep{zhou2022large}
& 33.8 & 4.80 
& 30.3 & 5.95 
& 4.82 & 9.75 \\
\midrule
\textbf{Response Refinement} & & & & & & \\
\quad \textit{Self-Refine}~\citep{madaan2023self}
& 38.1 & 4.82 
& 31.2 & 5.94 
& 7.32 & 11.81 \\
\midrule
\textbf{Data Selection} & & & & & & \\
\quad \textit{Filtering (Original)} 
& 34.1 & 4.76 
& 28.0 & 5.90 
& 4.70 & 9.00 \\
\quad \textit{Filtering (Refined)} 
& 31.6 & 4.57 
& 31.2 & 6.03 
& 4.93 & 10.09 \\
\quad \textit{R.I.P}~\citep{yu2025rip} 
& 29.7 & 4.71
& 23.9 & 6.03
& 4.55 & 8.63 \\
\midrule
\rowcolor{gray!10} \cellcolor{white} \textbf{Refined Dataset}
& \textbf{47.8} & \textbf{5.38} 
& \textbf{35.8} & \textbf{6.39} 
& \textbf{8.86} & \textbf{18.78} \\
\bottomrule
\end{tabular}
}
\vspace*{-0.5em}
\caption{Results on MT-Bench, Evol-Instruct, and AlpacaEval across dataset improvement strategies. WR: win rate; Score: average score.}
\label{tab:refinement_comparison}
\vspace*{-1em}
\end{table}
\section{Experiments}~\label{sec:4_experiments}
To test whether improving instruction quality improves preference learning, we evaluate instruction refinement in both offline and online settings.
The offline setting refines a fixed preference dataset, whereas the online setting refines instructions during iterative on-policy data generation.

\subsection{Offline Preference Learning}~\label{subsec:4_1_offline}
\noindent\textbf{Experimental Setup.}
We conduct offline experiments on 32K instructions randomly sampled from UltraFeedback~\citep{cui2024ultrafeedback}.
For each instruction, we generate on-policy responses with the target model, apply three iterations of instruction refinement, and then construct the final preference dataset.
We evaluate two target backbones, LLaMA3-8B~\citep{grattafiori2024llama} and Mistral-7B~\citep{jiang2023clip}, initialized from SFT checkpoints, and train them with DPO~\citep{rafailov2023direct} and SimPO~\citep{meng2024simpo}.

We use GPT-4o for feedback and refinement.
ArmoRM~\citep{ArmoRM} is used both to score responses during refinement and to assign final preference labels.
The refinement threshold is set to \(\tau=0.13\), corresponding to approximately the top 30\% of the initial on-policy reward distribution.
We evaluate on MT-Bench~\citep{zheng2023judging}, Evol-Instruct~\citep{xu2023wizardlm}, and AlpacaEval~\citep{alpaca_eval}; implementation details are provided in Appendix~\ref{C_training_details}.

\begin{table*}[ht]
\centering
\footnotesize
\resizebox{0.9\linewidth}{!}{
\begin{tabular}{l|cc|cc|cc}
\toprule
\multirow{2}{*}{\textbf{Method}} 
& \multicolumn{2}{c|}{\textbf{MT-Bench}} 
& \multicolumn{2}{c|}{\textbf{Evol-Instruct}} 
& \multicolumn{2}{c}{\textbf{AlpacaEval 2.0}} \\
& Win Rate & Score & Win Rate & Score & Win Rate & LC Win Rate \\
\midrule SFT & 27.8 & 4.61 & 23.4 & 5.8 & 3.95 & 7.49\\ \midrule DPO & 39.4 & 5.12 & 37.4 & 6.05 & 7.73 & 12.33 \\
\midrule
SPA \textit{(iter 3)} 
& 50.6 & 5.48 & \textbf{55.0} & 6.91 & 22.89 & 21.52 \\
\hspace{0.8em}+ \textit{w/ instruction refinement} 
& \textbf{55.9} (\textcolor{green}{↑} 5.3\%p) & \textbf{5.49} (\textcolor{green}{↑} 0.01) 
& \textbf{55.0} (--) & \textbf{7.24} (\textcolor{green}{↑} 0.33) 
& \textbf{23.60} (\textcolor{green}{↑} 0.71\%p) & \textbf{21.81} (\textcolor{green}{↑} 0.29\%p) \\
\midrule
Reward Model \textit{(iter 3)} 
& \textbf{47.8} & \textbf{5.40} 
& \textbf{47.0} & 6.39 
& \textbf{11.78} & 19.92 \\
\hspace{0.8em}+ \textit{w/ instruction refinement} 
& 47.8 (--) & 5.33 (\textcolor{red}{↓} 0.07) 
& 44.3 (\textcolor{red}{↓} 2.7\%p) & \textbf{6.61} (\textcolor{green}{↑} 0.22) 
& 11.51 (\textcolor{red}{↓} 0.26\%p) & \textbf{20.62} (\textcolor{green}{↑} 0.69\%p) \\
\midrule
Implicit Reward (iter 3) 
& 51.9 & \textbf{5.68} 
& 52.3 & 6.78 
& 18.40 & 24.62 \\
\hspace{0.8em}+ w/ instruction refinement 
& \textbf{54.4} (\textcolor{green}{↑} 2.5\%p) & 5.49 (\textcolor{red}{↓} 0.19) 
& \textbf{57.6} (\textcolor{green}{↑} 5.3\%p) & \textbf{6.91} (\textcolor{green}{↑} 0.13) 
& \textbf{25.53} (\textcolor{green}{↑} 7.14\%p) & \textbf{25.03} (\textcolor{green}{↑} 0.41\%p) \\
\bottomrule
\end{tabular}}
\vspace*{-0.5em}
\caption{
Evaluation results on MT-Bench, Evol-Instruct, and AlpacaEval 2.0 for LLaMA-3-8B models trained with SPA (iter 3), reward model annotation, and implicit reward annotation.
\textbf{Bold} indicates the better result within each method, and arrows indicate the change compared to the non-refined counterpart.}
\label{tab:online_iter3_extended}
\end{table*}

\noindent\textbf{Experimental Results.}
Table~\ref{tab:1_offline} reports offline results for DPO and SimPO on both backbones.
Refined datasets yield broad gains over original datasets across objectives, backbones, and benchmarks, with especially clear improvements on AlpacaEval LC win rate.
These results indicate that improving instruction quality can strengthen preference data without changing the preference-learning objective.
We further demonstrate on Llama-3-8B feedback and refiner and a different reward model~\footnote{\textit{OpenAssistant/reward-model-deberta-v3large-v2}} setting in Appendices~\ref{sec:a_module} and ~\ref{a_7_alt_rm}, respectively.

\noindent\textbf{Baseline Comparisons.}
We compare our methods to other baseline methods: Prompt rewriting includes CoT prompting~\citep{wei2022chain} and generic paraphrasing~\citep{zhou2022large}; response refinement uses Self-Refine~\citep{madaan2023self}; and data selection includes reward-based filtering and R.I.P.~\citep{yu2025rip}, which select examples using response-level signals such as reward, length, and reward gap.
These baselines modify the instruction surface, generated responses, or dataset composition, whereas our method refines weak instructions to improve preference data without discarding examples.

As shown in Table~\ref{tab:refinement_comparison}, instruction refinement achieves the strongest overall performance among all compared interventions.
The gains over CoT prompting and paraphrasing indicate that generic rewriting is not sufficient, while gains over Self-Refine and data selection show that repairing weak instructions provides benefits beyond response-level refinement or filtering.
These results support instruction quality as a distinct axis of preference data improvement.

\noindent\textbf{Refined Dataset Analysis.}
We examine whether refinement changes task intent or reduces task difficulty.
Human evaluation shows that 88.4\% of refined instructions preserve the core task intent (see Appendix~\ref{a_8_human_eval}).
We also compare Instruction-Following Difficulty (IFD) scores~\cite{li2024quantity} and find only a small change, with a slight increase after refinement.
This suggests that the gains are unlikely to come simply from making instructions easier to follow (Appendix~\ref{a_9_ifd}).

\begin{table*}[!ht]
\centering
\resizebox{0.9\textwidth}{!}{
\begin{tabular}{c|cc|cc|cc|cc|cc|cc}
\toprule
\multirow{2}{*}{\textbf{Model}} 
& \multicolumn{6}{c|}{\textbf{DPO}} 
& \multicolumn{6}{c}{\textbf{SimPO}} \\
\cmidrule(lr){2-7} \cmidrule(lr){8-13}
& \multicolumn{2}{c|}{MT-Bench} 
& \multicolumn{2}{c|}{Evol-Instruct} 
& \multicolumn{2}{c|}{AlpacaEval} 
& \multicolumn{2}{c|}{MT-Bench} 
& \multicolumn{2}{c|}{Evol-Instruct} 
& \multicolumn{2}{c}{AlpacaEval} \\
& Pairwise & Single 
& Pairwise & Single 
& WR & \textbf{LC WR} 
& Pairwise & Single 
& Pairwise & Single 
& WR & \textbf{LC WR} \\
\midrule
SFT
& 27.8\% & 4.61 
& 23.4\% & 5.80 
& 3.55 & 7.15 
& 27.8\% & 4.61 
& 23.4\% & 5.80 
& 3.55 & 7.15 \\
\midrule
Original-OSSAT 
& \textbf{41.9\%} & 5.11 
& 29.8\% & \textbf{6.32} 
& 7.90 & 11.71 
& 37.5\% & \textbf{5.26} 
& 26.4\% & 6.41 
& \textbf{8.75} & 12.32 \\
\rowcolor{gray!10}
Refined-OSSAT 
& 40.9\% & \textbf{5.24} 
& \textbf{36.5\%} & 6.26 
& \textbf{7.99} & \textbf{13.86} 
& \textbf{44.7\%} & 5.07 
& \textbf{32.3\%} & \textbf{6.48} 
& 8.37 & \textbf{13.31} \\
\bottomrule
\end{tabular}
}
\vspace*{-0.5em}
\caption{Cross-reward-model evaluation results on MT-Bench, Evol-Instruct, and AlpacaEval. For MT-Bench and Evol-Instruct, each cell reports the pairwise win rate and single-answer score, while for AlpacaEval, it reports the win rate and length-controlled win rate. The highest values in each column are highlighted in \textbf{bold}.}
\label{tab:ossat_offline}
\end{table*}

\subsection{Online Preference Learning}\label{subsec:4_2_offline}
\noindent\textbf{Experimental Setups.}
In online preference learning, the model generates preference data and is then updated on those data, allowing the policy and data distribution to co-evolve.
This setting tests whether instruction quality continues to limit newly generated data as the model improves over iterations.

We follow online preference learning pipelines from SPA~\citep{kim2025spread} and iterative DPO, where labels are obtained from model-generated responses using either implicit rewards or an external reward model; SPA further smooths small-gap preference signals.
We integrate instruction refinement by applying one refinement step to all instructions before each response-generation round, without threshold-based selection.
The underlying online preference-learning algorithm is otherwise unchanged, isolating the effect of improving instruction quality during iterative data generation.

Following the SPA setup, we use UltraFeedback and initialize the policy from a LLaMA3-8B SFT checkpoint.
Training starts with DPO on 3.3\% gold-labeled data, followed by three online iterations with 8K, 20K, and 30K on-policy samples.
PairRM~\citep{jiang2023llm} is used for reward-model preference annotation.

\noindent\textbf{Experimental Results.}
Table~\ref{tab:online_iter3_extended} shows that instruction refinement improves most online preference learning variants relative to training with original instructions.
The gains are strongest for SPA at iteration 3, where all benchmarks improve, and AlpacaEval LC win rate improves consistently across SPA, reward-model, and implicit-reward variants.
These gains are notable because online learning continually updates the policy that generates new data; nevertheless, weak instructions remain a bottleneck in the evolving data-generation process.

The gains are broad but not uniform across all metrics.
MT-Bench score slightly decreases for reward-model and implicit-reward variants despite improved or maintained win rates.
A case study in Appendix~\ref{app:degradation_cases} suggests that such drops often occur in tasks requiring strict output formats, indicating that online instruction refinement remains task-dependent.
Additional intermediate-iteration results are provided in Appendix~\ref{a_online}.

\begin{figure*}[!t]
    \centering
    \includegraphics[width=\textwidth]{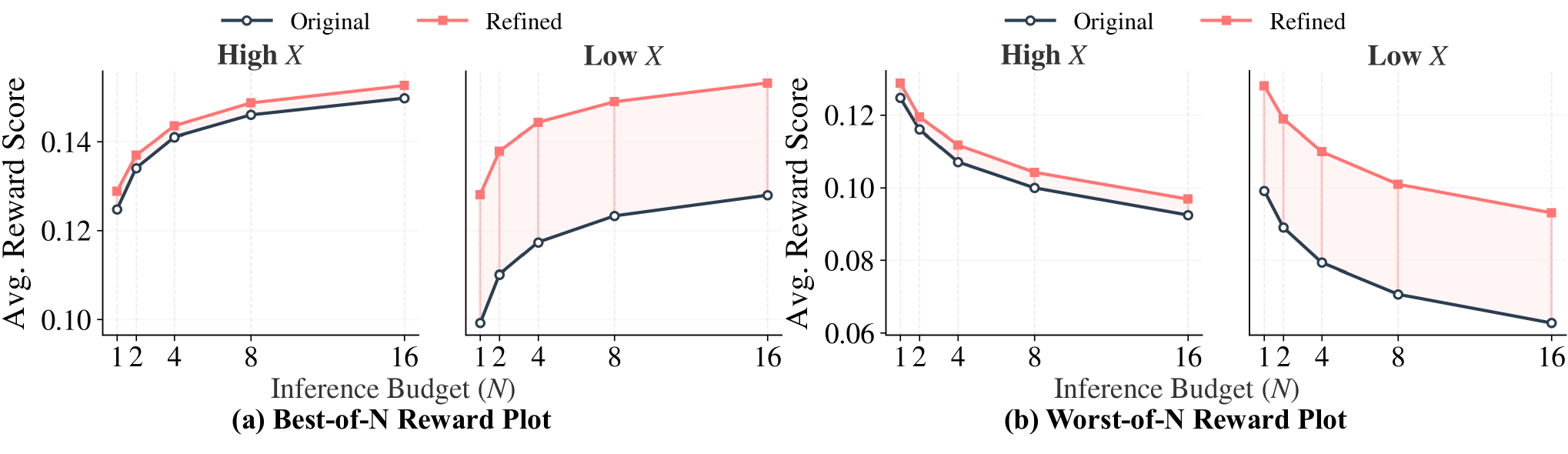}
    \caption{
        Best-of-$N$ and Worst-of-$N$ analysis showing that instruction refinement raises the response quality ceiling across high and low instruction quality groups, with more pronounced improvements for lower-quality instructions.
    }
    \label{fig:4_refined_bon}
    \vspace{-1em}
\end{figure*}
\begin{figure}[!t]
    \centering
    \includegraphics[width=\linewidth]{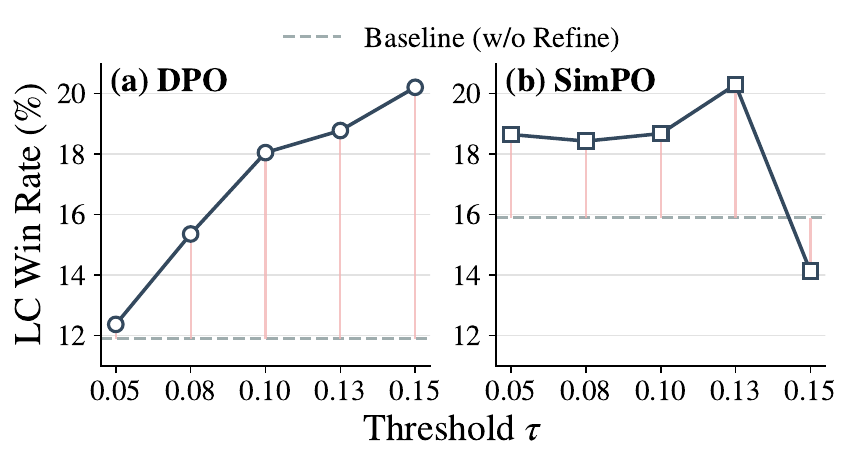}
    \caption{
        Sensitivity analysis of the minimum-reward threshold $\tau$ for instruction refinement. Refinement consistently improves over the non-refined baseline, with optimal performance at moderate thresholds.
    }
    \label{fig:7_sensitivity}
    \vspace{-1em}
\end{figure}

\section{Discussion} 
\subsection{Cross-Reward Model Experiment}~\label{subsec:5_2_cross_rm}
In Section~\ref{subsec:4_1_offline}, we use ArmoRM both to select instructions for refinement and to construct preference labels.
To examine whether the observed gains are specific to this reward model, we conduct a cross-reward-model experiment.
Specifically, we use ArmoRM to select instructions for refinement, while using OSSAT Reward Model~\citep{openassistant_deberta_rm} as reward model to assign preference labels.
Using the resulting datasets, we train models with the same DPO and SimPO pipelines and compare \textit{Original-OSSAT} with \textit{Refined-OSSAT}.

The results in Table~\ref{tab:ossat_offline} show that Refined-OSSAT generally performs better than Original-OSSAT under the OSSAT-RM annotation setting.
While the gains are not uniform across all metrics, the overall trend suggests that refined-instruction data remains effective even when preference labels are assigned by a different reward model.
This suggests that the benefit of our method is not specific to the bias of a single reward model.

\noindent\textbf{Reward Model Human Evaluation.}
To further examine reward-model dependency, we conducted a small-scale human evaluation on 30 instruction--response-pair tuples, evenly split between examples selected and not selected for refinement.
Each tuple was annotated by three evaluators, with the final decision determined by majority vote.

The reward-model-based refinement decision agreed with the human majority judgment in 76.7\% of cases.
For preference labeling, the reward-model label agreed with the human majority choice in 71.4\% of cases.
These results suggest that the reward-model signals used for both instruction selection and preference-label construction are reasonably aligned with human judgments.
Nevertheless, the imperfect agreement with human judgments may be attributed in part to known reward-model biases toward factors such as response length, style, or surface-level fluency~\citep{bu2025beyond}.
Therefore, in our pipeline, the reward-model objective should be interpreted as a proxy for human preference rather than a direct substitute.
Details of the annotation protocol and prompt are provided in Appendix~\ref{a_9_rm_human_eval}.

\subsection{Bo$N$ \& Wo$N$ Analysis}
Figure~\ref{fig:4_refined_bon} shows how response quality changes after instruction refinement under a fixed sampling budget using Best-of-$N$ (Bo$N$) and Worst-of-$N$ (Wo$N$) evaluation.
Across two instruction quality groups (\textit{High-X, Low-X}), instruction refinement consistently shifts both Bo$N$ and Wo$N$ curves upward, indicating a higher achievable reward ceiling as well as an improved reward floor.
The effect is most pronounced for the \textit{Low-X} group.
Before refinement, low-quality instructions exhibit a lower average reward even with increased sampling, reflecting a tight upper bound imposed by poor instruction quality.
After refinement, the Bo$N$ and Wo$N$ curves for \textit{Low-X} rise significantly, suggesting that instruction improvement can alleviate limitations induced by low-quality instructions.
Overall, these results suggest that improving instruction quality can raise achievable response quality while also improving the lower tail of the response-quality distribution under a fixed sampling budget.

\subsection{Threshold Analysis}~\label{subsec:5_4_threshold_analysis}
In the offline setting, we select instructions for refinement by thresholding the minimum response reward.
A higher threshold makes refinement stricter, filtering more low-quality instructions but increasing risk of over-constrained data.
To analyze sensitivity to the threshold \(\tau\), we vary \(\tau \in \{0.05, 0.08, 0.10, 0.13, 0.15\}\) under the same setup as Section~\ref{subsec:4_1_offline}.
Figure~\ref{fig:7_sensitivity} reports AlpacaEval LC win rates for DPO and SimPO across thresholds.

Across most values of \(\tau\), instruction refinement consistently outperforms the baseline without refinement, indicating robust improvements in alignment performance.
Performance generally increases as the threshold becomes more selective, suggesting that refining lower-reward instructions yields more informative preference data.
DPO performs best around \(\tau = 0.13\) (refining \(\sim\)30\% of instructions) and remains robust at higher thresholds.
In contrast, SimPO degrades at higher thresholds, likely due to overly aggressive refinement that reduces response diversity and weakens preference signals.
Overall, instruction refinement is broadly robust, with moderate thresholds yielding the most consistent gains.
However, the degradation of SimPO at higher thresholds highlights a practical limitation, as overly aggressive refinement can reduce response diversity and weaken preference signals.
Additional sensitivity analyses are provided in Appendix~\ref{sensitivity_app}.

\subsection{Rubric Effect Analysis}~\label{subsec:5_3_rubric_ablation}
\noindent\textbf{Rubric Ablation.}
Although our method uses GPT-4o to refine instructions, the observed gains may partly come from general rewriting by a strong foundation model.
To disentangle these effects, we conduct a rubric ablation in which GPT-4o refines the instructions without access to rubric-based feedback.
Both settings refine the entire dataset once for a controlled comparison.
In this setting, GPT-4o rewrites the instructions using the same guidance but without access to rubric-based feedback.
This design isolates the contribution of rubrics while controlling for the refine model and prompt.
\begin{table}[t]
\centering
\resizebox{\columnwidth}{!}{
\begin{tabular}{l|cc|cc|cc}
\toprule
\textbf{Method}
& \multicolumn{2}{c|}{MT-Bench} 
& \multicolumn{2}{c|}{Evol-Instruct} 
& \multicolumn{2}{c}{AlpacaEval} \\
& \textbf{WR} & \textbf{Score} 
& \textbf{WR} & \textbf{Score} 
& \textbf{WR} & \textbf{LC WR} \\
\midrule
\textbf{SFT}
& 27.8 & 4.61 
& 23.4 & 5.80 
& 3.55 & 7.15 \\
\textbf{DPO} (Original Dataset)
& 39.7 & 4.91 
& 33.3 & 6.15 
& 5.91 & 11.90 \\
\midrule
\textbf{Refined Dataset} & & & & & & \\
\textit{w/o Rubric-Based Prompt}
& \textbf{47.8} & 5.52 
& 36.2 & 6.54 
& 8.60 & 16.23 \\
\rowcolor{gray!10} \cellcolor{white}
\textit{w/ Rubric-Based Prompt}
& 47.5 & \textbf{5.60} 
& \textbf{41.7} & \textbf{6.70} 
& \textbf{9.92} & \textbf{19.54} \\
\bottomrule
\end{tabular}
}
\caption{Rubric ablation results across evaluation benchmarks. WR: win rate; Score: average score.}
\label{tab:rubric_feedback_ablation}
\end{table}
\begin{table}[t]
\centering
\resizebox{\columnwidth}{!}{
\begin{tabular}{l|cc|cc}
\toprule
\textbf{$\Delta$ Rubric Score}
& \multicolumn{2}{c|}{\textbf{w/o Rubric}} 
& \multicolumn{2}{c}{\textbf{w/ Rubric}} \\
& \textbf{$\Delta$ Avg.} & \textbf{$\Delta$ Min.}
& \textbf{$\Delta$ Avg.} & \textbf{$\Delta$ Min.} \\
\midrule
Clarity
& 0.048 & 0.047
& \textbf{0.064} & \textbf{0.077} \\
Specificity
& -0.041 & -0.029
& \textbf{-0.008} & \textbf{0.015} \\
Completeness
& \textbf{-0.028} & -0.017
& -0.032 & \textbf{-0.005} \\
Safety
& 0.062 & 0.057
& \textbf{0.091} & \textbf{0.091} \\
Answerability
& -0.002 & 0.009
& \textbf{0.056} & \textbf{0.078} \\
Conciseness
& \textbf{0.097} & \textbf{0.087}
& 0.089 & 0.086 \\
Format Consistency
& \textbf{0.016} & \textbf{0.018}
& 0.003 & 0.014 \\
\bottomrule
\end{tabular}
}
\caption{Pearson correlation between rubric score gains and average/minimum reward score gains, with and without rubric-based feedback.}
\label{tab:rubric_feedback_reward_deltas}
\vspace*{-1em}
\end{table}
As shown in Table~\ref{tab:rubric_feedback_ablation}, rubric-based feedback yields improvements over rubric-free refinement on all metrics except the MT-Bench win rate.
This indicates that rubric-guided refinement provides robust gains beyond general instruction rewriting by GPT-4o, suggesting that our seven rubric dimensions help produce instructions that are more beneficial for preference learning.

\noindent\textbf{Rubric-Reward Correlation.}
We further examine whether rubric-based feedback better aligns instruction-level improvements with downstream reward gains than rubric-free feedback.
For each rubric dimension, we compute the correlation between rubric score gains and response reward gains from the original to the refined instruction.
We consider two reward gain metrics: average reward gain and minimum reward gain.
Since each instruction has two responses, average reward is their mean score, while minimum reward is the lower score.
This dimension-level analysis identifies which rubric improvements are most closely associated with downstream reward gains.

Table~\ref{tab:rubric_feedback_reward_deltas} presents the results.
Rubric-based refinement shows stronger positive correlations for safety, answerability, and clarity than rubric-free refinement, suggesting that rubric-based feedback makes improvements in these dimensions more aligned with downstream reward gains.
By contrast, specificity and completeness show near-zero or negative correlations with reward gains.
This suggests that these dimensions are less directly associated with reward increases, although they may capture quality aspects not fully reflected by the reward model.
Overall, the results indicate that rubric-based feedback better aligns instruction  improvements with response reward gains.

\noindent\textbf{Additional Analyses.}
We provide additional analyses to further examine the validity and generality of instruction refinement.
First, we conduct a human evaluation to assess whether refinement preserves the original task intent.
The results show that most refined instructions retain the core task, suggesting that the observed gains are not primarily driven by task drift or distributional replacement (Appendix~\ref{a_8_human_eval}).
We also evaluate refinement on a medical-domain preference dataset and observe improvements on the QA-style medical benchmark (UltraMedical-Preference~\cite{zhang2024ultramedical}), indicating that instruction refinement can also benefit preference learning in domain-specific settings (Appendix~\ref{a_10_ultramed}).

\section{Conclusion}
We show that instruction quality can limit achievable response quality and, consequently, preference learning performance.
Through Best-of-$N$ analysis, we provide evidence that low-quality instructions impose a ceiling on response quality under a fixed sampling budget.
We further show that improving instruction quality strengthens preference signals without discarding data, in both offline and online settings.
Across multiple algorithms and benchmarks, instruction refinement consistently improves preference data quality and alignment performance.
Overall, these results suggest that instruction quality is an important factor in determining the effectiveness of preference learning, beyond response quality and optimization objectives.
\section*{Limitation}
Our study has several limitations.
First, our approach relies on a reward model to identify and refine low-quality instructions, and its effectiveness may depend on the model’s calibration and biases.
Second, aggressive refinement can degrade performance in some settings, indicating sensitivity to threshold selection.
Third, instruction refinement introduces additional computational cost due to iterative feedback and rewriting, although it is generally more cost-efficient than collecting new preference data.
Finally, our study is limited to text-only language models, and extending this approach to multimodal settings remains an important direction for future work.

\bibliography{custom}

\clearpage
\appendix

\section{Motivating Analysis Details}\label{subsec:a_1_motivating}

This matters for preference learning because both tails affect pairwise data informativeness.
Let \(F_x(t)=\Pr_{Y\sim\mu_x}[R(x,Y)\le t]\) be the response-quality distribution induced by instruction \(x\), where \(R(x,y)\) is a reward-based proxy for response quality.
With \(K\) sampled candidates, the probability of obtaining a high-quality response above \(\tau_h\) is \(1-F_x(\tau_h)^K\), while the probability of obtaining a trivial response below \(\tau_l\) is \(1-(1-F_x(\tau_l))^K\).
Thus, improving the upper tail increases coverage of strong chosen candidates, while reducing lower-tail mass decreases degenerate rejected candidates that can form overly easy preference pairs.
This motivates instruction refinement as an upstream intervention before candidate response generation.

We introduce our motivation by presenting evidence from Best-of-$N$ (Bo$N$) and Worst-of-$N$ (Wo$N$) experiments, showing that instruction quality constrains both the upper and lower tails of response quality.
Bo$N$ selects the highest-reward response among $N$ generated candidates, providing an estimate of the maximum attainable response quality, while Wo$N$ selects the lowest-reward response, characterizing the quality floor of responses induced by an instruction.
We partition 1.5K sampled instructions from the UltraFeedback dataset into three quality groups (\textit{High}, \textit{Mid}, and \textit{Low}) based on our instruction quality assessment, which scores each instruction on seven criteria and uses the average score to assign a quality tier (refer to Section~\ref{subsec:3_2_eval_ref_instruction}), and generate multiple responses per instruction using LLaMA3-8B~\cite{grattafiori2024llama}.

As shown in Figure~\ref{fig:1_plot_sample}(a), Bo$N$ reward improves with increasing $N$, but a substantial gap persists between high- and low-quality instruction groups even at large sampling budgets.
At $N=16$, high-quality instructions achieve a mean Bo$N$ reward of 0.150, compared to 0.128 for low-quality instructions, yielding a gap of 0.0218 ($p<0.0001$).
This suggests that instruction quality constrains the observed upper tail of response quality under the sampled candidate distribution.
Figure~\ref{fig:1_plot_sample}(b) further shows that the same trend holds for Wo$N$: high-quality instructions also yield higher worst-case rewards than low-quality instructions.
At $N=16$, high-quality instructions achieve a mean Wo$N$ reward of 0.092, compared to 0.063 for low-quality instructions, corresponding to a gap of 0.0297 ($p<0.0001$).
Together, these results suggest that high-quality instructions improve both the upper and lower tails of the sampled response-quality distribution.

To interpret these results, let \(\mu_x(y)=\mu(y\mid x)\) denote the candidate response distribution induced by instruction \(x\).
For each sampled response \(Y_i\sim \mu_x\), let \(R^*(x,Y_i)\) denote its latent response quality.
Given \(N\) sampled candidates \(Y_1,\ldots,Y_N\sim\mu_x\), the Bo\(N\) and Wo\(N\) experiments can be viewed as measuring the upper and lower order statistics of this quality distribution:
\begin{equation}
\begin{aligned}
    R_{\max}^{(N)}(x) &= \max_{i\in[N]} R^*(x,Y_i),\\
    R_{\min}^{(N)}(x) &= \min_{i\in[N]} R^*(x,Y_i).
\end{aligned}
\end{equation}
The observed Bo\(N\) gap indicates that high-quality instructions induce candidate distributions with better upper tails.
The observed Wo\(N\) gap further indicates that they also raise the lower tail, reducing the likelihood of very poor sampled responses.

Preference data are constructed from these sampled candidates by selecting a chosen response \(Y^+\) and a rejected response \(Y^-\).
Let
\begin{equation}
    r^+ = R^*(x,Y^+), \qquad r^- = R^*(x,Y^-)
\end{equation}
denote the latent qualities of the chosen and rejected responses.

\paragraph{Coverage Enables Direct Gradients.}
When an instruction is low quality, the induced candidate distribution may rarely contain strong responses.
For a high-quality threshold \(\tau\), suppose that low-quality instructions assign negligible probability mass to high-quality responses:
\begin{equation}
    \Pr_{Y\sim\mu_{x_{\mathrm{low}}}}
    \left[
    R^*(x_{\mathrm{low}},Y)\ge \tau
    \right]
    \approx 0.
\end{equation}
Since \(Y^+\) is selected from a finite set of candidates sampled from \(\mu_{x_{\mathrm{low}}}\), the probability of obtaining a high-quality chosen response is also small:
\begin{equation}
    \Pr
    \left[
    R^*(x_{\mathrm{low}},Y^+)\ge \tau
    \mid x_{\mathrm{low}}
    \right]
    \approx 0.
\end{equation}

For a preference tuple \((x,y^+,y^-)\), the DPO gradient~\citep{rafailov2023direct} can be written as
\begin{equation}
\nabla_\theta \mathcal{L}_{\mathrm{DPO}}
=
-\beta
\mathbb{E}_{(x,y^+,y^-)\sim \mathcal{D}}
\left[
\sigma(-u_\theta)d_\theta
\right],
\end{equation}
where
\begin{equation}
u_\theta
=
\beta\left[
\log\frac{\pi_\theta(y^+\mid x)}{\pi_{\mathrm{ref}}(y^+\mid x)}
-
\log\frac{\pi_\theta(y^-\mid x)}{\pi_{\mathrm{ref}}(y^-\mid x)}
\right],
\end{equation}
and
\begin{equation}
d_\theta
=
\nabla_\theta \log \pi_\theta(y^+\mid x)
-
\nabla_\theta \log \pi_\theta(y^-\mid x).
\end{equation}
The first term in \(d_\theta\) increases the likelihood of the chosen response.
Therefore, if high-quality chosen responses are rarely observed, DPO receives little direct gradient signal for increasing the likelihood of high-quality responses.~\citep{pan2025what}
In contrast, high-quality instructions make such chosen responses more likely to appear in preference data, yielding more useful positive updates.

\paragraph{Higher Rejected Floors Yield More Informative Pairs.}
Instruction quality also affects the rejected response because \(Y^-\) is selected from candidates sampled from the same response distribution \(\mu_x\).
A higher-quality lower tail increases the typical value of \(r^-\), thereby reducing the quality gap \(r^+ - r^-\) between chosen and rejected responses.
When a large quality gap between \(r^+\) and \(r^-\) is reflected in the learned preference margin \(u_\theta\), we have
\begin{equation}
    u_\theta \gg 0
    \quad\Rightarrow\quad
    \sigma(-u_\theta)\approx 0.
\end{equation}
Such pairs have a large quality gap between the chosen and rejected responses, making them easy to distinguish and less informative for gradient updates~\citep{houliston2024uncertainty}.
In contrast, when the quality gap \(r^+ - r^-\) is smaller and reflected in \(u_\theta\), the resulting margin can be smaller~\citep{lee2025dataset}:
\[
u_\theta \not\gg 0
\quad\Rightarrow\quad
\sigma(-u_\theta)\not\approx 0.
\]
Thus, improving instruction quality can provide more direct positive updates toward high-quality chosen responses while maintaining informative gradients from pairs with smaller chosen--rejected quality gaps.

\section{Experimental Details}\label{sec:b_exp_details}
\subsection{Instruction Evaluation Rubric Details}\label{appendix:rubric_design}
We survey prior studies on active preference learning and preference dataset selection to identify which aspects of instructions degrade preference data quality, and derive seven instruction refinement rubrics from this analysis.

\begin{itemize}
    \item \textbf{Clarity} assesses whether the instruction clearly specifies what the model is expected to do without ambiguity.
    Prior work shows that ambiguous instructions can lead models to misidentify the intended action, motivating consistency in interpretation~\citep{wang2025ask}.
    \item \textbf{Specificity} evaluates whether the instruction provides sufficiently concrete requirements, as vague or underspecified instructions have been shown to degrade model performance~\citep{zhou2023instruction, kim2025detail}.
    Together, these criteria encourage instructions to state requirements explicitly and precisely.
    \item \textbf{Completeness} measures whether all necessary information and constraints to answer an instruction are provided.
    Prior work shows that missing context or incomplete task definitions are associated with failures independent of model capability, highlighting issues in instruction design~\citep{wang2025ask, liang2022holistic}.
    Providing a complete specification is therefore essential for reliable response generation.
    \item \textbf{Safety} evaluates whether the instruction may elicit harmful or unsafe outputs.
    Previous studies report that unsafe instructions can induce harmful behaviors or generate dangerous information~\citep{bai2022constitutional, perez2022red}.
    We explicitly assess instruction safety to ensure refinement remains within acceptable response boundaries.
    \item \textbf{Answerability} examines whether the instruction admits a well-defined and reasonably answerable response.
    Several data collection studies identify answerability as a core criterion for high-quality instruction data, noting that unanswerable or ill-posed instructions lead to failures regardless of model strength~\citep{mahmud2025maple, zhu2025fanno}.
    We therefore treat answerability as a first-class criterion in instruction evaluation.
    \item \textbf{Conciseness} assesses whether the instruction avoids unnecessary information that may distract the model from core requirements.
    Prior work shows that excessive constraints or irrelevant details can impair generation quality~\citep{liu2023g}.
    This criterion ensures that instructions convey only information essential for the task.
    \item \textbf{Format Consistency} evaluates whether the desired output format is clearly specified and consistently illustrated, for example through few-shot examples.
    Inconsistent or underspecified formats have been shown to cause instruction-following failures~\citep{zhou2023instruction,liu2025reife}.
\end{itemize}

For each rubric dimension, we use a 5-point Likert-style scale.
This choice is motivated by prior human and AI evaluation practices in natural language generation.
In particular, recent AI-feedback annotation pipelines such as UltraFeedback~\citep{cui2024ultrafeedback} use 1--5 scalar scores with detailed score documentation for each aspect to reduce variability and subjectivity in annotation standards.
Furthermore, because our refinement procedure is methodologically inspired by Self-Refine~\citep{madaan2023self}, we follow its use of bounded, multi-dimensional 1--5 scoring prompts with textual feedback.

\subsection{Feedback and Refinement Prompt}~\label{A_6_prompt}
As described in section~\ref{subsec:3_2_eval_ref_instruction}, we include the prompts used as input to the LLM-as-a-judge for feedback generation and instruction refinement in Figure ~\ref{feedback} and ~\ref{refine}.
These prompts are adapted from prior Self-Refine~\cite{madaan2023self}, with newly defined rubric criteria for instruction-level refinement.
The prompts are designed to robustly elicit both rubric-based scores and natural language feedback, and we strictly constrain the output format and provide few-shot examples to ensure consistent and accurate refinement.
Especially, to preserve the original intent during refinement, we explicitly impose the constraints to preserve the intent of the original on the instructions.

\subsection{Benchmarks}\label{subsec:a_3_benchmarks}
For both main experimental settings, LLM alignment is evaluated using three benchmarks---MT-Bench~\cite{zheng2023judging}, Evol-Instruct~\cite{xu2023wizardlm}, and AlpacaEval 2.0~\cite{alpaca_eval}.
These benchmarks adopt an \texttt{LLM-as-a-judge} evaluation paradigm and report Win Rate, defined as the proportion of cases in which the target model’s response is preferred over that of a comparison model.
For MT-Bench and Evol-Instruct, we additionally report single score evaluations that quantify absolute response quality.
For AlpacaEval, we report both the standard Win Rate and the length-controlled Win Rate to mitigate biases arising from response length.

MT-Bench~\cite{zheng2023judging} and Evol-Instruct~\cite{xu2023wizardlm} are alignment benchmarks designed to evaluate instruction-following capabilities under different levels of interaction and complexity.
MT-Bench focuses on multi-turn dialogue and instruction execution, assessing a model’s ability to maintain coherent and natural conversations across two-turn interactions.
In contrast, Evol-Instruct addresses the bias of existing benchmarks toward relatively simple instructions by automatically generating instruction data with diverse difficulty levels using LLMs, resulting in a more challenging evaluation set.

Both benchmarks adopt the same automatic evaluation protocol based on the \texttt{llm-judge} framework~\cite{zheng2023judging}, with \texttt{GPT-4o-2024-11-20} serving as the judge model. For each instruction, responses generated by the target model are compared against those produced by \texttt{GPT-3.5-turbo} to assign win/lose/tie outcomes, enabling the computation of win rates. In addition, the judge produces scalar quality scores for individual responses, and we report the averaged single score for each benchmark.

AlpacaEval~\cite{alpaca_eval} is an LLM-based benchmark that evaluates general instruction-following performance. Each example consists of an instruction paired with two responses: one generated by the target model and the other by a reference model, \texttt{GPT-4-1106-preview}. Judging is conducted by \texttt{GPT-4o-mini-2024-07-18}, which compares the two responses and assigns a win/lose/tie label according to how well each fulfills the instruction. We report the win rate of the target model relative to the reference model.

\section{Training Details}~\label{C_training_details}
In this section, we provide the training setup used in our experiments for reproducibility, including the basic configuration and hyperparameter settings.
In the main experiments, both DPO and SimPO training were conducted, and we implemented both using the LLaMA-Factory~\cite{zheng2024llamafactory} codebase.
\subsection{DPO and SimPO Training Details}
\paragraph{Offline Settings.}
In the offline setting, we conduct experiments using hyperparameters that achieved the best performance identified through random hyperparameter search.

\begin{table*}[!ht]
\centering
\resizebox{0.95\textwidth}{!}{
\begin{tabular}{c|c|cc|cc|cc|cc|cc|cc}
\toprule
\multicolumn{2}{c|}{\textbf{Method}}& \multicolumn{6}{c|}{DPO} & \multicolumn{6}{c}{SimPO} \\
\midrule
\multirow{2}{*}{\textbf{Backbone}} & \multirow{2}{*}{\textbf{Data Type}} & \multicolumn{2}{c|}{MT-Bench} & \multicolumn{2}{c|}{Evol-Instruct} &  \multicolumn{2}{c|}{AlpacaEval} & \multicolumn{2}{c|}{MT-Bench} & \multicolumn{2}{c|}{Evol-Instruct} & \multicolumn{2}{c}{AlpacaEval} \\
& & WR & Score & WR & Score & WR & \textbf{LC WR} & WR & Score & WR & Score & WR & \textbf{LC WR} \\
\midrule
\multirow{2}{*}{Llama3-8b}
& SFT & 27.8 & 4.61 & 23.4 & 5.80 & 3.95 & 7.49 & 27.8 & 4.61 & 23.4 & 5.80 & 3.95 & 7.49 \\
\cmidrule(lr){2-14}
& Original & 45.0 & \textbf{5.57} & \textbf{40.6} & 6.60 & 8.74 & 17.59 & 45.6 & \textbf{5.45} & 38.1 & 6.63 & 7.08 & 14.92 \\
\rowcolor{gray!10} \cellcolor{white} & \cellcolor{white} Refined & \textbf{51.6} & 5.53 & 36.0 & \textbf{6.63} & \textbf{9.26} & \textbf{19.00} & \textbf{48.8} & 5.39 & \textbf{41.1} & \textbf{6.64} & \textbf{8.87} & \textbf{17.07} \\
\bottomrule
\end{tabular}
}
\caption{Evaluation results of prompt refinement experiment with the Feedback \& Refine module, conducted under the same experimental setting as Table~\ref{tab:1_offline}. The highest values for each backbone are highlighted in \textbf{bold}.}
\label{tab:1_offline_module}
\end{table*}
\label{sec:a_module_exp}

For DPO, the learning rate is selected from [$2e-6, 1e-6, 5e-7, 6e-7$] and $\beta$ is chosen from [$0.01, 0.05, 0.1$]. For SimPO, hyperparameter tuning is performed over learning rates [$3e-6, 2e-6, 1e-6, 3e-7, 5e-7, 6e-7$], $\beta$ values [$1, 2, 2.5$], and $\gamma$ values [$1.0, 1.6$]. In LLaMA3-8B SimPO, learning rate range is [$3e-6, 2e-6, 1e-6, 3e-7, 5e-7, 6e-7$], and in Mistral-7B SimPO, [$3e-7, 5e-7, 2e-6, 1e-5, 2e-5$]. The optimal hyperparameters were selected separately for each model backbone.
In our experiments, for LLaMA-3-8B, DPO training was conducted with $\beta$ = 0.01 and a learning rate of $1e-6$, while SimPO used $\beta$ = 1, a learning rate of $2e-6$, and $\gamma$ = 1.0.
For Mistral-7B, DPO was trained with $\beta$ = 0.01 and a learning rate of $2e-6$. SimPO employed $\beta$ = 1, a learning rate of $2e-6$, and $\gamma$ = 1.0.

The LLaMA-3-8B model was trained for 3 epochs, while the Mistral-7B model was trained for 1 epoch. Both models were trained with a batch size of 16.
All experiments were conducted using two NVIDIA H200 GPUs.

\paragraph{Online Settings.}
In the online setting, we followed the SPA~\cite{kim2025spread} experiment configuration and used the same hyperparameters. 
In this experiment, we employed a LLaMA-3-8B backbone and performed one epoch of DPO training per iteration with a learning rate of $1e-5$ and $\beta$ = 0.01.
We also utilize two NVIDIA H200 GPUs in online setting.

\subsection{Feedback and Refinement Module Training.}
In feedback and refinement module training, we adopted supervised fine-tuning in LLaMA-3-8B model. 
We utilized GPT-generated 16k feedback data and refinement data, which produce in UltraFeedback offline experiment in Section~\ref{subsec:4_1_offline}.
SFT was conducted with learning rate of $1e-4$, and we train model for 3 epoch with batch size 16. 

\section{Additional Experiments}\label{c_additional_exp}
\subsection{Feedback \& Refinement Module Experiment}~\label{sec:a_module}
\noindent
To demonstrate that refinement is possible by replacing the foundation model with a module trained via supervised fine-tuning (SFT), we conducted module experiments under the same settings as in Section~\ref{subsec:4_1_offline}.
\paragraph{Experimental Setups.}
As for the fine-tuning LLaMA3-8B, we use another unseen instructions in the UltraFeedback and collect the feedback and refinement results of them from GPT-4o.
Similarly, we performed three iteration of instruction refinement. 
Models were trained on the dataset before refinement and on the refined dataset, and their performance was subsequently compared.
\paragraph{Experimental Results.}
Table~\ref{tab:1_offline_module} shows the offline preference learning results, where utilized a feedback module and refiner module trained via SFT instead of \texttt{GPT-4o}.
Consistent with previous results, table also demonstrates that refined data leads to superior training performance across nearly all benchmarks.
This demonstrates that a trained refiner module can effectively polish data as an alternative to a foundation model.
\begin{table*}[!ht]
\centering
\resizebox{0.95\textwidth}{!}{
\begin{tabular}{c|c|cc|cc|cc|cc|cc|cc}
\toprule
\multicolumn{2}{c|}{\textbf{Method}}& \multicolumn{6}{c|}{DPO} & \multicolumn{6}{c}{SimPO} \\
\midrule
\multirow{2}{*}{\textbf{Backbone}} & \multirow{2}{*}{\textbf{Data Type}} & \multicolumn{2}{c|}{MT-Bench} & \multicolumn{2}{c|}{Evol-Instruct} & \multicolumn{2}{c|}{AlpacaEval} & \multicolumn{2}{c|}{MT-Bench} & \multicolumn{2}{c|}{Evol-Instruct} & \multicolumn{2}{c}{AlpacaEval} \\
& & WR & Score & WR & Score & WR & \textbf{LC WR} & WR & Score & WR & Score & WR & \textbf{LC WR} \\
\midrule
\multirow{3}{*}{Llama3-8b}
& SFT & 0.278 & 4.61 & 0.234 & 5.80 & 3.55 & 7.15 & 0.278 & 4.61 & 0.234 & 5.80 & 3.55 & 7.15 \\
\cmidrule(lr){2-14}
& Original & \textbf{0.466} & \textbf{5.47} & 0.374 & 6.39 & \textbf{9.43} & 16.56 & \textbf{0.466} & 5.23 & 0.401 & \textbf{6.56} & 3.61 & 8.07 \\
\rowcolor{gray!10} \cellcolor{white}
& Refined & 0.453 & 5.22 & \textbf{0.383} & \textbf{6.47} & 9.38 & \textbf{17.20} & \textbf{0.466} & \textbf{5.30} & \textbf{0.436} & 6.49 & \textbf{6.75} & \textbf{13.84} \\
\bottomrule
\end{tabular}
}
\caption{Results with an alternative reward model on MT-Bench, Evol-Instruct, and AlpacaEval. For MT-Bench and Evol-Instruct, conducted under the same experimental setting as Table~\ref{tab:1_offline}. The highest values for each backbone are highlighted in \textbf{bold}.}
\label{tab:alt_rm_results}
\end{table*}
\begin{table*}[ht]
\centering
\resizebox{0.95\textwidth}{!}{
\begin{tabular}{l|cccccc|c|c|c}
\toprule
\textbf{Method}
& \multicolumn{6}{c|}{MMLU-med}
& \multicolumn{1}{c|}{MedQA}
& \multicolumn{1}{c|}{MedMCQA} 
& \multicolumn{1}{c}{PubMedQA}\\
& \makecell{\textbf{Clinical}\\\textbf{Knowledge}}
& \makecell{\textbf{Medical}\\\textbf{Genetics}}
& \textbf{Anatomy}
& \makecell{\textbf{Professional}\\\textbf{Medicine}}
& \makecell{\textbf{College}\\\textbf{Biology}}
& \makecell{\textbf{College}\\\textbf{Medicine}}
& \textbf{Acc.}
& \textbf{Acc.}
& \textbf{Acc.} \\
\midrule
\textbf{SFT}
& 0.592 & 0.710 & 0.556 & 0.592 & 0.618 & 0.578 
& 0.515 & 0.486 & 0.022 \\
\midrule
\textbf{ DPO} (Original)
& 0.660 & 0.620 & 0.593 & \textbf{0.607} & \textbf{0.681} & 0.566 
& 0.534 & 0.496 & 0.196 \\
\rowcolor{gray!10} \cellcolor{white}
\textbf{Refined}
& \textbf{0.694} & \textbf{0.680} & \textbf{0.593} & 0.588 & \textbf{0.681} & \textbf{0.578}
& \textbf{0.537} & \textbf{0.502} & \textbf{0.254} \\
\bottomrule
\end{tabular}
}
\caption{Medical-domain evaluation results on MMLU-med, MedQA, and PubMedQA. The highest values in each column are highlighted in \textbf{bold}.}
\label{tab:medical_eval}
\vspace*{-1em}
\end{table*}
\subsection{Alternative Reward Model Experiment}~\label{a_7_alt_rm}
We have demonstrated that our framework works with two different reward models---ArmoRM in the offline setting and PairRM in the online setting. 
To further verify that our approach does not depend on a specific reward model, we conduct additional experiments with an alternative reward model, \textit{OpenAssistant/reward-model-deberta-v3-large-v2}, a general-purpose reward model trained on diverse datasets such as WebGPT and HH-RLHF.
All experimental settings follow those described in Section~\ref{subsec:4_1_offline}, and we conduct DPO and SimPO on LLaMA-3-8B.

Table~\ref{tab:alt_rm_results} reports the results.
Overall, refining instructions yields improvements on multiple benchmarks and settings under the alternative reward model.
In particular, clearer improvements are observed on Evol-Instruct and AlpacaEval, while some MT-Bench metrics show smaller gains or slight decreases.
These results suggest that the proposed approach is not tied to a single reward model configuration, while also indicating that the magnitude and consistency of the improvements are affected by the reward signal.

\subsection{Alternative Dataset Experiment}~\label{a_10_ultramed}
Since our main experiments are conducted on UltraFeedback, the observed performance gains could be specific to the data characteristics of UltraFeedback.
To examine this possibility, we conduct an additional DPO experiment with LLaMA-3-8B using TsinghuaC3I/UltraMedical-Preference, a medical-domain preference dataset.
We use 33K examples corresponding to 30\% of the full 110K UltraMedical-Preference dataset.
Following the setup in Section~\ref{sec:4_experiments}, we train LLaMA-3-8B-SFT with DPO on original and refined dataset.
For instruction refinement, we use GPT-4o with the same configuration as in our main experiments, applying a refinement threshold of 0.13 for two iterations.

The results in Table~\ref{tab:medical_eval} show an overall positive trend from instruction refinement, while the effect varies across MMLU-med.
The improvements are particularly clear on QA-style benchmarks, where the refined model achieves higher scores on MedQA, MedMCQA, and PubMedQA.
These results suggest that our pipeline can also provide benefits on preference datasets from different domains.

\section{Additional Analysis}\label{e_additional_analysis}
\begin{figure}[!t]
    \centering
    \includegraphics[width=0.9\linewidth]{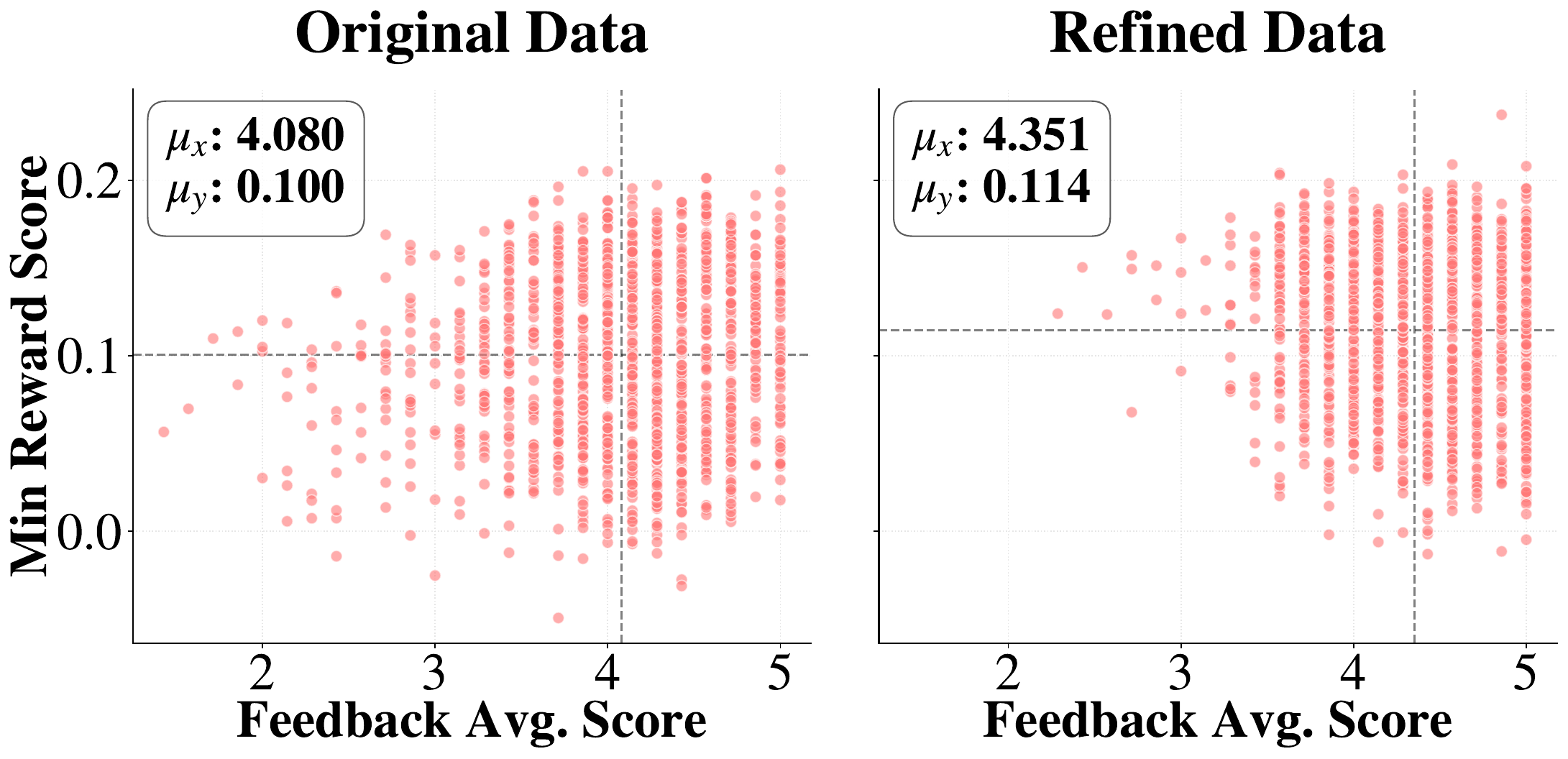}
    \caption{
        Distribution of feedback scores and minimum response rewards before and after instruction refinement, showing consistent improvements in both instruction quality and lower-bound response quality.
    }
    \label{fig:8_refined_scatter}
    \vspace{-1em}
\end{figure}
\subsection{Instruction–Response Quality Analysis}

To analyze the relationship between instruction quality and response quality, we present a scatter plot in Figure~\ref{fig:8_refined_scatter} that compares the distributions of feedback scores and minimum reward scores before and after instruction refinement.
Each point represents an instruction, with the horizontal axis denoting the average rubric-based feedback score and the vertical axis indicating the minimum reward score across sampled responses.

After refinement, we observe a clear upward shift along both axes.
The average minimum reward score rises from $\mu_y = 0.100$ to $\mu_y = 0.114$, which corresponds to 11.65\% of the effective reward range of ArmoRM scores on UltraFeedback.

\begin{table*}[ht]
\centering
\footnotesize
\resizebox{\linewidth}{!}{
\begin{tabular}{l|cc|cc|cc}
\toprule
\multirow{2}{*}{\textbf{Method}} & \multicolumn{2}{c|}{\textbf{MT-Bench}} & \multicolumn{2}{c|}{\textbf{Evol-Instruct}} & \multicolumn{2}{c}{\textbf{AlpacaEval 2.0}} \\
& Win Rate & Score & Win Rate & Score & Win Rate & LC Win Rate \\
\midrule
SFT & 27.8 & 4.61 & 23.4 & 5.8 & 3.95 & 7.49\\
\midrule
DPO & 39.4 & 5.12 & 37.4 & 6.05 & 7.73 & 12.33 \\
\midrule
SPA \textit{(iter 2)} & 53.1 & 5.56 & \textbf{56.0} & 6.46 & 19.14 & 20.27 \\
\hspace{0.8em}+ \textit{w/ instruction refinement} & \textbf{54.7} (\textcolor{green}{↑} 1.6\%p) & \textbf{5.59} (\textcolor{green}{↑} 0.03) & 55.7 (↓ 0.3\%p) & \textbf{6.89} (\textcolor{green}{↑} 0.43) & \textbf{20.80} (\textcolor{green}{↑} 1.66\%p)& \textbf{20.82} (\textcolor{green}{↑} 0.55\%p) \\
\midrule
Reward Model \textit{(iter 2)} & 43.1 & \textbf{5.19} & 40.8 & \textbf{6.41} & \textbf{9.19} & 15.49 \\
\hspace{0.8em}+ \textit{w/ instruction refinement} & \textbf{45.3} (\textcolor{green}{↑} 2.2\%p) & 5.18 (\textcolor{red}{↓} 0.01) & \textbf{42.9} (\textcolor{green}{↑} 2.1\%p) & 6.33 (\textcolor{red}{↓} 0.08) & 9.01	(\textcolor{red}{↓} 0.18\%p) & \textbf{15.62} (\textcolor{green}{↑} 0.14\%p) \\
\midrule
Implicit Reward \textit{(iter 2)} & \textbf{53.1} & \textbf{5.49} & 48.4 & 6.54 & 14.27 & 21.38\\
\hspace{0.8em}+ \textit{w/ instruction refinement} & 50.9 (\textcolor{red}{↓} 2.2\%p) & 5.36 (\textcolor{red}{↓} 0.13) & \textbf{55.5} (\textcolor{green}{↑} 7.1\%p) & \textbf{6.61} (\textcolor{green}{↑} 0.07) & \textbf{21.90} (\textcolor{green}{↑} 7.63\%p) & \textbf{22.84} (\textcolor{green}{↑} 1.46\%p) \\
\bottomrule
\end{tabular}}
\caption{\label{tab:2_online_iter2}
Evaluation results on MT-Bench, Evol-Instruct, and AlpacaEval 2.0 for LLaMA-3-8B models trained with SPA (iter 2).
Bold indicates the better result within each method, and arrows indicate the change compared to the non-refined counterpart.}
\end{table*}

\begin{figure*}[!t]
    \centering
    \includegraphics[width=\textwidth]{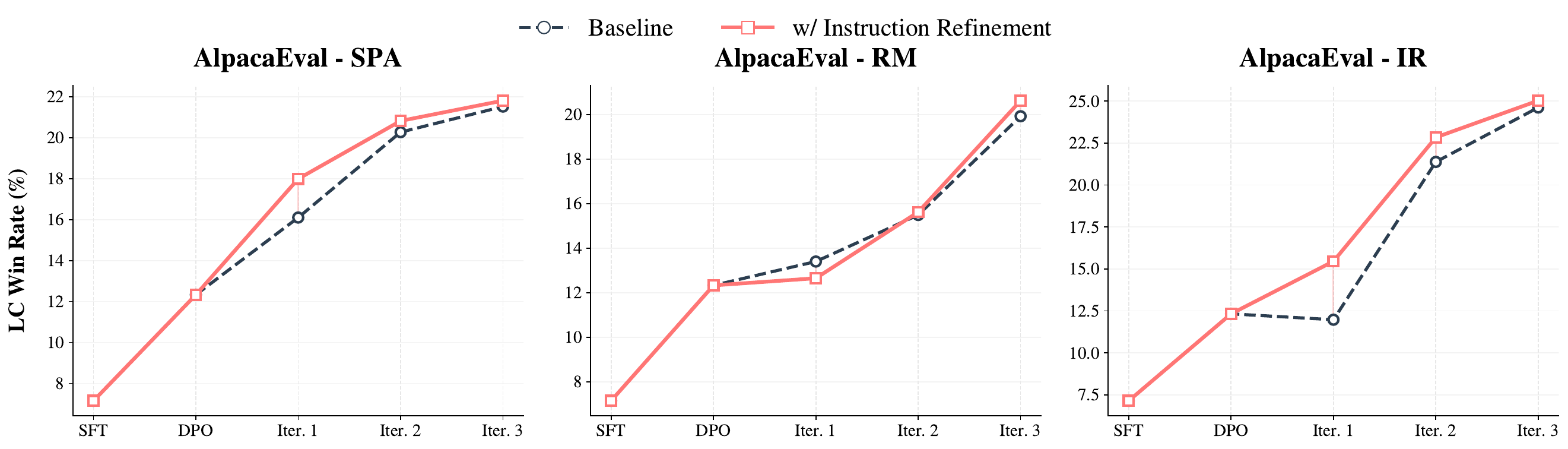}
    \caption{
        AlpacaEval LC win rate result per iteration of online preference learning. Model capability increases as train iteration repeat, showing major increase in SPA + prompt refinement.
    }
    \label{fig:3_spa_graph}
    \vspace{-1em}
\end{figure*}

This observation suggests that refinement reduces the prevalence of low-quality generations.
In other words, refined instructions are more likely to elicit consistently high-quality responses, rather than exhibiting unstable generation behavior.
This distributional shift helps explain the observed improvements in both offline and online preference learning performance. 
These patterns suggest that high-quality instructions are associated with reliable response quality and strong preference signals.

\subsection{Online Experiment Analysis}~\label{a_online}
To provide a more detailed analysis of the SPA experiments presented in the main paper (Section~\ref{subsec:4_2_offline}), this section reports the results at iteration 2 and presents performance gain across iterations, illustrating how model performance evolves as iterative preference learning progresses.

Table~\ref{tab:2_online_iter2} summarizes the results of online preference learning with two iterations of SPA, comparing baseline training with and without instruction refinement.
Across most settings, incorporating instruction refinement consistently improves or maintains performance over the non-refined counterparts, demonstrating its effectiveness as a complementary mechanism in online preference learning.
Notably, SPA with instruction refinement yields consistent gains across most benchmarks, while reward-model-based and implicit-reward-based settings also benefit from refinement, particularly on Evol-Instruct and AlpacaEval.
These results indicate that refining queries improves the quality of on-policy preference data, leading to more informative learning signals during online training.

Figure~\ref{fig:3_spa_graph} further illustrates performance trends across training iterations.
In most settings—SPA and implicit-reward-based training—instruction refinement accelerates performance improvements as training progresses.
These results show that instruction refinement enhances the effectiveness of online preference learning by improving data quality throughout training, rather than only at initialization.


\begin{figure*}[!t]
    \centering
    \includegraphics[width=\textwidth]{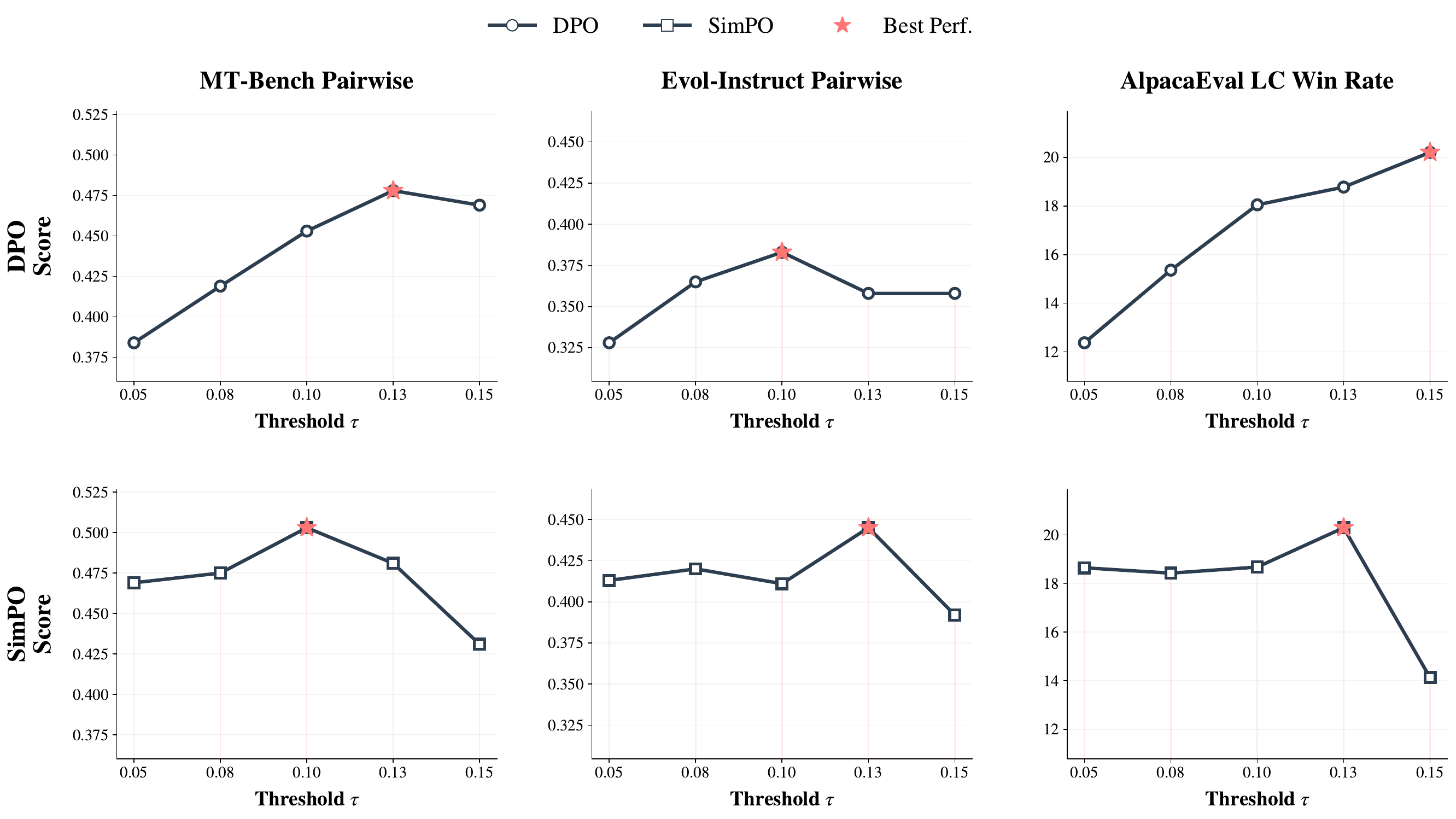}
    \caption{
        $\tau$ Sensitivity Analysis Result in MT-Bench~\cite{zheng2023judging}, Evol-Instruct~\cite{xu2023wizardlm}, and AlpacaEval 2.0~\cite{alpaca_eval}.
    }
    \label{fig:9_sensitivity_5}
\end{figure*}
\begin{table*}[!ht]
\centering
\resizebox{0.8\textwidth}{!}{
\begin{tabular}{c|c|ccc|c|ccc}
\toprule
\multirow{2}{*}{\textbf{Dataset}} 
& \multicolumn{4}{c|}{\textbf{MT-Bench}} 
& \multicolumn{4}{c}{\textbf{Evol-Instruct}} \\
\cmidrule(lr){2-5} \cmidrule(lr){6-9}
& Overall & Low-IFD & Mid-IFD & High-IFD 
& Overall & Low-IFD & Mid-IFD & High-IFD \\
\midrule
Original 
& 4.913 & 4.302 & 5.056 & 4.925 
& 6.147 & \textbf{6.342} & 6.208 & 6.411 \\
\rowcolor{gray!10}
Refined 
& \textbf{5.381} & \textbf{5.037} & \textbf{5.717} & \textbf{5.396} 
& \textbf{6.390} & 6.205 & \textbf{6.417} & \textbf{6.548} \\
\bottomrule
\end{tabular}
}
\caption{Evaluation results across different IFD difficulty levels on MT-Bench~\citep{zheng2023judging} and Evol-Instruct~\citep{xu2023wizardlm}. Each cell reports the average score. The highest value in each column is highlighted in \textbf{bold}.}
\label{tab:ifd_analysis}
\vspace{-1em}
\end{table*}
\subsection{Threshold Sensitivity Analysis}\label{sensitivity_app}
Figure~\ref{fig:9_sensitivity_5} reports the sensitivity of instruction refinement performance to the minimum-reward threshold~$\tau$ across different evaluation benchmarks for both DPO and SimPO.
Across MT-Bench, Evol-Instruct, and AlpacaEval, moderate threshold values consistently lead to improved performance compared to the non-refined baseline, indicating that selectively refining low-quality queries effectively enhances preference learning.
In particular, both methods achieve their best or near-best performance around $\tau = 0.1$–$0.13$, suggesting that filtering and refining approximately the lower portion of queries strikes a balance between removing unstable instructions and preserving informative preference signals.
When the threshold is set too high (e.g., $\tau = 0.15$), performance sometimes degrades, implying that excessive refinement may over-constrain instructions or reduce preference signal effectivity.
\begin{table*}[ht]
\centering
\small
\begin{tabular}{l|p{5.5cm}|p{6.5cm}}
\toprule
\textbf{Category} & \textbf{Original instruction} & \textbf{Refined instruction} \\
\midrule
Major Shift 
& How to use corkscrew to chop the vegetables? 
& Ensure safety by securely holding a knife to chop carrots, onions, and bell peppers on a stable cutting board. \\
\midrule
Minor Shift 
& Create a sentence using five synonyms 
& Use five synonyms for ``happy'' in a single sentence. \\
\bottomrule
\end{tabular}
\caption{Examples of major and minor instruction shifts between original and refined instructions.}
\label{tab:instruction_shift_examples}
\end{table*}
\begin{figure*}[!t]
    \centering
    \includegraphics[width=0.9\textwidth]{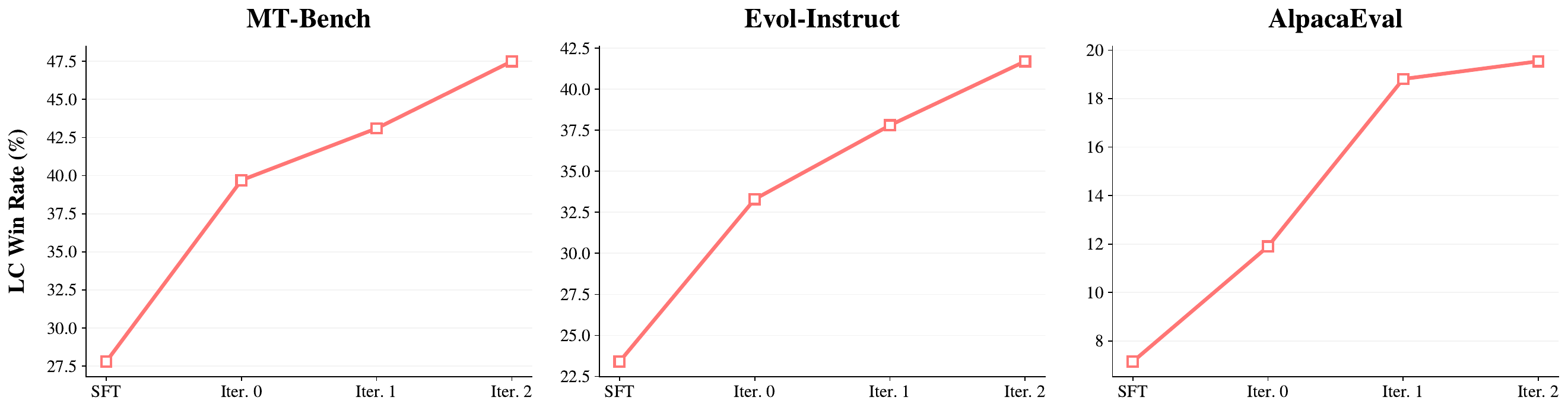}
    \caption{Performance across refinement iterations. Win rate generally improves from SFT to later refinement iterations across MT-Bench, Evol-Instruct, and AlpacaEval.}
    \label{fig:10_iter_plot}
    \vspace{-1em}
\end{figure*}
Overall, these results confirm that instruction refinement is robust to threshold choice within a reasonable range, while highlighting the importance of moderate filtering for stable and effective preference optimization.

\subsection{IFD Score Analysis}~\label{a_9_ifd}
\textbf{IFD Score Experiment.} 
Task difficulty and instruction quality can be closely related, making it non-trivial to determine whether improved downstream performance stems from better instruction quality or from a reduction in task difficulty.
In particular, if instruction refinement reduces task difficulty, the resulting performance gains may be confounded with difficulty reduction rather than reflecting improvements in instruction quality alone.

To examine whether our refinement process reduces task difficulty, we conducted an additional analysis using Instruction-Following Difficulty (IFD) scores~\citep{li2024quantity}.
We use IFD as a proxy for instruction-following difficulty, with higher scores indicating greater difficulty under the model used to compute IFD.
For an instruction--answer pair $(Q, A)$, the IFD score is defined as:
\begin{equation}
\mathrm{IFD}_{\theta}(Q, A)
=
\frac{s_{\theta}(A \mid Q)}{s_{\theta}(A)}
\end{equation}

We computed IFD scores for each instance before and after instruction refinement using the Llama-3-Base-8B-SFT model and compared the resulting paired scores.
After refinement, the mean IFD score increased from 0.6423 to 0.6823, yielding a mean paired difference of +0.0406.
In addition, 54.38\% of the instances had higher IFD scores after refinement.
The difference was limited in magnitude, and the increase in IFD suggests that the refined instructions were not easier but marginally more difficult to follow.
Therefore, the observed improvement in model performance is unlikely to be attributable to reduced task difficulty in the refined data.

\textbf{Benchmark Evaluation.} 
To further examine whether the downstream gains are driven by easier data, we conducted an additional analysis on the evaluation benchmark.
If the observed improvements were mainly driven by easier training data, the refined model would be expected to show degraded performance on evaluation prompts with higher IFD scores.
We therefore grouped the evaluation prompts into \textit{low-, middle-,} and \textit{high-IFD} groups based on their IFD scores and compared model performance within each group.

The results in Table~\ref{tab:ifd_analysis} show that the gains from instruction-quality refinement were not confined to low-difficulty evaluation prompts.
On both Evol-Instruct and MT-Bench, the refined model outperformed the original model across most IFD-based difficulty groups, including prompts with relatively high IFD scores.
This suggests that instruction-quality refinement remains effective not only for easier prompts but also for more difficult instructions.
Therefore, model performance gains are unlikely to be explained solely by task difficulty.

\subsection{Effect of Refinement Iterations on Performance}
To examine how performance changes with repeated refinement, we evaluate models trained on instructions refined over multiple iterations.
We report the win rate for MT-Bench and Evol-Instruct, and the length-controlled win rate for AlpacaEval.
As shown in Figure~\ref{fig:10_iter_plot}, performance generally improves from SFT to later refinement iterations across MT-Bench, Evol-Instruct, and AlpacaEval.
The gains are most pronounced in the early refinement stages, while later iterations show smaller improvements or signs of saturation.
This suggests that iterative refinement can further improve downstream performance, but its marginal benefit diminishes after a small number of iterations.

\section{Human Evaluation}\label{e_human_eval}
To further strengthen the robustness of the refinement process, we conducted two small-scale human evaluations.
The first evaluation was designed to verify whether task drift occurred during refinement.
In this evaluation, human annotators directly compared the instructions before and after refinement and assessed whether the underlying task was preserved.
The second evaluation was conducted to further examine the alignment between the reward model and human preferences.
Human evaluators were asked to review the instruction and the corresponding model responses, assess their preference between the responses, and determine whether instruction refinement was necessary.

\subsection{Task/Distribution Shift Human Evaluation}~\label{a_8_human_eval}
To verify that intent preservation is achieved in practice, we conduct a human evaluation on task/intent drift.
We quantitatively evaluate intent and task preservation between the original and refined instructions.
The evaluation is conducted on 60 randomly selected instructions, with each instruction assessed by three human annotators (12 annotators in total, each evaluating 20 instructions). Annotators are shown the original and refined instructions in random order and asked to judge whether they are semantically equivalent, exhibit a minor shift, or exhibit a major shift.
\begin{itemize}
    \item \textbf{\textit{Same intent}} : The core task and output type remain unchanged, with only changes in expression.
    \item \textbf{\textit{Minor shift}} : The core task is preserved, but some constraints or output formats are modified.
    \item \textbf{\textit{Major shift}} : The task type or underlying objective is altered.
\end{itemize}
Differences in instruction length, wording, or style are not considered intent drift.
The results show 66.7\% \textbf{\textit{Same Intent}}, 21.7\% \textbf{\textit{Minor Shift}}, 8.3\% \textbf{\textit{Major Shift}}, and 3.3\% Tie.
Overall, 88.4\% of samples preserve the core task (\textbf{\textit{Same Intent}} + \textbf{\textit{Minor Shift}}), indicating that intent preservation is largely maintained despite occasional shifts.

We also conduct case studies of minor and major intent shifts in table ~\ref{tab:instruction_shift_examples}.
In \textbf{\textit{major shift }}example, instruction asks how to cut vegetables with a corkscrew involves using a tool in a dangerous and unintended way.
Following safety and answerability criteria, the refiner shifts the intent to providing safety guidance and requesting the use of appropriate tools. 
In \textbf{\textit{minor shift}} example, instruction requests to generate sentences using five synonyms is unanswerable without a target word.
The refiner adds a reference word (e.g., happy), introducing a minor shift while preserving the main task and input/output format.
These cases illustrate that even when intent shifts occur, they can lead to improved preference data quality.
The instructions given to human evaluators are provided in Figure~\ref{instruction}.

\subsection{Reward Model Human Evaluation}\label{a_9_rm_human_eval}
To assess how well our reward-model-based pipeline aligns with human preferences, we conducted a small-scale human evaluation on 30 randomly sampled instruction--response-pair tuples.
The evaluation set consisted of 15 samples selected by our pipeline as requiring refinement and 15 samples not selected for refinement.
Each tuple was annotated by three human evaluators, and the final decision was determined by majority vote.
(9 annotators in total, each evaluating 10 instructions)

To approximate the conditions of our actual pipeline, annotators were asked to judge whether the instruction should be refined after inspecting the two associated responses.
If either response revealed an issue attributable to the instruction, annotators were instructed to mark the instruction as requiring refinement.
In addition, annotators assigned scores to the two responses and selected the better one, thereby producing a human chosen/rejected preference label.
The detailed annotation prompt is provided in Figure~\ref{rm_instruction}.

The reward-model-based refinement decision agreed with the human majority judgment in 76.7\% of cases.
For preference labeling, the reward-model label agreed with the human majority choice in 71.4\% of cases.
These results indicate that both instruction selection and preference labeling are reasonably aligned with human judgments.
However, reward models can still be biased by factors such as response length, style, and surface-level fluency.
Thus, the reward model in our pipeline should be viewed as a reasonably accurate proxy for human preference, rather than a perfect substitute.
\begin{table}[!t]
    \centering
    \footnotesize
    \begin{tabular}{p{0.9\linewidth}}
        \toprule
        \textbf{Original instruction:} How long is the needle inside the monkey's ear? \\
        \midrule
        \textbf{Original Response 1:} The needle inside the monkey's ear is 2cm long. \\
        \textbf{Original Response 2:} The needle is 2.5cm long and it is inserted into the monkey's ear. [...] \\
        \midrule
        \textbf{Feedback (Evaluator):} Clarity: 1/5, Specificity: 1/5, Completeness: 1/5, Safety: 3/5, Answerability: 1/5, Conciseness: 3/5, FormatConsistency: 4/5. \textit{"Unclear context, lacks details, unanswerable without additional information."} \\
        \midrule
        \midrule
        \textbf{Refined instruction:} Provide the typical length of needles used in veterinary procedures for monkeys' ears, specifying the context and purpose of the procedure. \\
        \midrule
        \textbf{Refined Response 1:} The typical length ranges from \textbf{1 to 1.5 inches}. These needles are usually used for ear canal biopsies or ear canal washes to diagnose [...] \\
        \textbf{Refined Response 2:} The typical length is \textbf{1 inch (2.5 cm)}. This needle length is used for blood collection, injection, or fluid removal [...] \\
        \bottomrule
    \end{tabular}
    \caption{Case study of instruction refinement.}
    \label{tab:monkey_needle_case}
    \vspace{-1em}
\end{table}

\section{Case Study}
\subsection{Refinement Case Study}~\label{a_7_case}
Table~\ref{tab:monkey_needle_case} presents a qualitative case study illustrating the effect of instruction refinement.
The original instruction is ambiguous and underspecified, asking for the length of a \textit{“needle inside the monkey’s ear”} without providing any context, which leads to arbitrary and inconsistent responses despite surface-level confidence.
The evaluator feedback identifies deficiencies in clarity and specificity, indicating that the instruction is fundamentally underspecified.
After refinement, the instruction is revised to specify a realistic veterinary context and purpose, resulting in responses that are more concrete, medically plausible, and mutually consistent.
This example illustrates how improving instruction quality can yield more reliable preference data.

\subsection{Degradation Case Study}\label{app:degradation_cases}
We further analyzed cases where the refined model receives lower evaluation scores than the original model.
These cases suggest that larger drops can occur in tasks where correctness or output format is critical, including probability reasoning and named entity extraction.
Table~\ref{tab:degradation_case_ner} shows a representative degradation case in named entity extraction (NER).
The task requires a JSON dictionary grouped into predefined entity types.
While the original model follows the requested format, the refined model adds non-requested content and includes invalid or misclassified entities.
This example illustrates that instruction refinement can occasionally be less stable for tasks requiring strict output formats.
This pattern is less evident in open-ended generation tasks, where score changes are more likely to reflect judge-score variance or response-style differences.

\begin{figure*}[t]
\centering
\fbox{
\begin{minipage}{0.95\textwidth}
\small\ttfamily
You are an evaluator. Given an Original instruction, evaluate the instruction using the criteria below. 
Follow these STRICT rules: \\
1. Output must start with exactly 'Evaluation:' on its own line. \\
2. You must include ALL 7 criteria in the following order: 
   Clarity, Specificity, Completeness, Safety, Answerability, Conciseness, FormatConsistency.
3. Each line must follow the format:
   * <Criterion>: <digit 1-5>/5 - <one concise note> \\
4. Do NOT add any text before or after the evaluation block. \\
\\
Output format: \\
Evaluation: \\
* Clarity: <1-5>/5 - <one-line note> \\
* Specificity: <1-5>/5 - <one-line note> \\
* Completeness: <1-5>/5 - <one-line note> \\
* Safety: <1-5>/5 - <one-line note> \\
* Answerability: <1-5>/5 - <one-line note> \\
* Conciseness: <1-5>/5 - <one-line note> \\
* FormatConsistency: <1-5>/5 - <one-line note> \\
--- \\
\#\#\# Few-shot Examples \\
Original instruction: \\
Write something about animals \\
Evaluation: \\
* Clarity: 2/5 - vague request \\
* Specificity: 2/5 - no length, no scope \\
* Completeness: 2/5 - missing output format \\
* Safety: 5/5 - safe \\
* Answerability: 4/5 - feasible but broad \\
* Conciseness: 3/5 - some redundancy \\
* FormatConsistency: 3/5 - unspecified output format \\
\\
\#\#\# \\
\\
Original instruction: \\
Make a paragraph using that language \\
Evaluation: \\
* Clarity: 2/5 - unclear what "that language" refers to \\
* Specificity: 2/5 - format is defined (paragraph) but content is vague \\
* Completeness: 2/5 - missing target language or subject \\
* Safety: 5/5 - safe request \\
* Answerability: 2/5 - partially answerable but underspecified \\
* Conciseness: 4/5 - concise but incomplete \\
* FormatConsistency: 3/5 - loosely consistent but ambiguous wording \\
\\
\#\#\# \\
\\
Original instruction: \\
\{instruction\} \\
\end{minipage}
}
\caption{Prompt used for Feedback Generation.}\label{feedback}
\end{figure*}
\begin{figure*}[t]
\centering
\fbox{
\begin{minipage}{0.95\textwidth}
\small\ttfamily
You are an instruction refiner. \\
Given an Original instruction and its Evaluation, rewrite the instruction so it is clear, specific, complete, safe, feasible, concise, and consistent. \\
\\
STRICT RULES: \\
1. Output ONLY the refined instruction text — no labels, no commentary, no examples. \\
2. The refined instruction must be executable directly by a model (not guidelines). \\
3. Preserve the intent of the original, but resolve ambiguity using the feedback. \\
4. Do not include the words "Refined instruction" in the output. \\
5. Write a single instruction only. \\
\\
Original instruction: 
\{instruction\} \\
Evaluation: 
\{feedback\} \\
\end{minipage}
}
\caption{Prompt used for Refinement Generation.}\label{refine}
\end{figure*}
\begin{figure*}[t]
\centering
\fbox{
\begin{minipage}{0.95\textwidth}
\small\ttfamily
You will compare two prompts (an original version and a revised version) and assess whether the revised prompt preserves the same underlying intent and task as the original. \\
This evaluation focuses on the core task the model is expected to perform, not on superficial differences in wording or style. \\

\#\#\# How to Evaluate \\
Read the two prompts (Version A and Version B) shown below and judge whether they require the model to perform the same task. \\

When making your decision, consider the following: \\

Whether the type of output expected from the model is the same \\
Whether the core operation the model must perform (e.g., translation, classification, reasoning, generation) is the same \\
Whether the goal or scope of the task has changed \\
Differences in phrasing, length, or politeness alone do not constitute intent drift. \\
If the model is expected to perform the same core action, select “Same intent.” \\

\#\#\# Options \\

Please choose one of the following: \\

Same intent : The two prompts differ only in wording or formatting; the core task and output type remain the same. \\
Minor shift : The core task is largely preserved, but some constraints or output format have changed. \\
Major shift : The type of task or the objective has changed. For example, the output type changes, or the prompt shifts from solving a problem to describing the task itself. \\

\#\#\# Notes \\
Do not base your judgment on prompt length, verbosity, or stylistic differences. \\
Focus on what the model is being asked to do. \\
If unsure, base your decision on whether the nature of the expected output is the same in both prompts. \\
\end{minipage}
}
\caption{Instructions provided to human evaluators for task shift evaluation.}\label{instruction}
\end{figure*}
\begin{figure*}[t]
\centering
\fbox{
\begin{minipage}{0.95\textwidth}
\small\ttfamily
You are a human evaluator. Your task is to evaluate two model responses to the same instruction. \\

For each evaluation item, follow this process: \\

Step 1. Score each response separately. \\

Give a score to Response 1. \\
Give a score to Response 2. \\
Judge based on how well each response follows the instruction, correctness, completeness, clarity, formatting, and overall usefulness. \\
Use the same scoring standard for both responses. \\

Step 2. Decide whether the instruction needs refinement. \\

If either responses receive low scores, determine that the instruction has issues, respond that instruction refinement is needed. \\
Check whether the cause is issues in the instruction’s clarity, specificity, completeness, safety, answerability, conciseness, or format consistency. \\

Step 3. Select <chosen> and <rejected>. \\

Choose the better response as chosen. \\
Mark the weaker response as rejected. \\
If both responses are equally poor or equally good, still choose the slightly better one. \\
\end{minipage}
}
\caption{Instructions provided to human evaluators for response evaluation and instruction refinement.}\label{rm_instruction}
\end{figure*}
\begin{table*}[t]
\centering
\begin{tabular}{p{16cm}}
\toprule
\textbf{Question:} Identify the named entities (people, organizations, locations) mentioned in the given news article. Please generate a JSON dictionary that lists the named entities in three separate groups based on their entity types. The key is the type of entity and the value is a list of strings. \\
\midrule
\textbf{Article:} Yesterday, Adamson Emerson, the CEO of Faraday, and Dieter Zetsche, the CEO of Daimler AG, announced plans to build a new Gigafactory in Berlin. The facility will be a joint venture between Faraday and Daimler, producing electric vehicles and battery packs for both companies, creating thousands of job opportunities in the region. Emerson and Zetsche stated that the strategic location of Berlin, coupled with its skilled workforce and strong infrastructure, makes it an ideal choice for expansion. The new Gigafactory aims to meet the growing demand for electric vehicles in Europe and contribute to a sustainable future. Volkswagen CEO Herbert Diess welcomed the news, saying greater collaboration will benefit the auto industry's transition to e-mobility. \\
\midrule
\end{tabular}

\vspace{0.5em}

\begin{tabular}{p{1.8cm}|p{11.5cm}|c}
\toprule
\textbf{Model} & \textbf{Model Response} & \textbf{Score} \\
\midrule
\textbf{Original} &
\begin{minipage}[t]{11.5cm}
\small
\begin{verbatim}
{
    "Person": ["Adamson Emerson", "Dieter Zetsche", "Herbert Diess"],
    "Organization": ["Faraday", "Daimler AG", "Volkswagen"],
    "Location": ["Berlin"]
}
\end{verbatim}
\end{minipage}
& 9 \\
\midrule
\textbf{Refined} &
\begin{minipage}[t]{11.5cm}
\small
\begin{verbatim}
Named Entities:
- Adamson Emerson (person)
- Dieter Zetsche (person)
- Faraday (organization)
- Daimler AG (organization)
- Berlin (location)
- Gigafactory (location)
- electric vehicles (product)
- battery packs (product)
- job opportunities (event)
- skilled workforce (attribute)
- strong infrastructure (attribute)
- sustainable future (goal)
- Herbert Diess (person)
- Volkswagen (organization)
- e-mobility (industry transition)

JSON Dictionary:
{
  "Person": ["Adamson Emerson", "Dieter Zetsche", "Herbert Diess"],
  "Organization": ["Faraday", "Daimler AG", "Volkswagen"],
  "Location": ["Berlin", "Gigafactory"]
}
\end{verbatim}
\end{minipage}
& 5 \\
\bottomrule
\end{tabular}
\caption{Representative degradation case in a strict-format named entity extraction task. The refined model adds non-requested content and misclassifies entities, leading to a lower evaluation score.}
\label{tab:degradation_case_ner}
\end{table*}

\end{document}